\documentclass[accepted]{article}

\usepackage{aistats2025}
\usepackage{amsmath,amssymb,amsfonts}
\usepackage{algorithmic}
\usepackage{amsmath,amssymb,amsfonts}
\usepackage{algorithm}
\usepackage{multirow}
\usepackage{siunitx}
\usepackage{algorithmic}
\usepackage{comment}
\usepackage{caption}
\usepackage{subcaption}
\usepackage{hyperref}
\usepackage[table,xcdraw]{xcolor}  

\usepackage{booktabs,
            makecell, 
            tabularx}
\usepackage{textcomp}
\usepackage{mathtools}
\usepackage[round]{natbib}

\begin{document}

%

%

\twocolumn[

\aistatstitle{DynaGraph: Dynamic Graph Learning for EHRs}


\aistatsauthor{ Munib Mesinovic\textsuperscript{1} \And Soheila Molaei\textsuperscript{1} \And  Peter Watkinson\textsuperscript{2} \And Tingting Zhu\textsuperscript{1}}

\aistatsaddress{} ]

\begin{abstract}
Learning from longitudinal electronic health records is limited if it does not capture the temporal trajectories of the patient's state in a clinical setting. Graph models allow us to capture the hidden dependencies of the multivariate time-series when the graphs are constructed in a similar dynamic manner. Previous dynamic graph models require a pre-defined and/or static graph structure, which is unknown in most cases, or they only capture the spatial relations between the features. Furthermore in healthcare, the interpretability of the model is an essential requirement to build trust with clinicians. In addition to previously proposed attention mechanisms, there has not been an interpretable dynamic graph framework for data from multivariate electronic health records (EHRs). Here, we propose DynaGraph, an end-to-end interpretable contrastive graph model that learns the dynamics of multivariate time-series EHRs as part of optimisation. We validate our model in four real-world clinical datasets, ranging from primary care to secondary care settings with broad demographics, in challenging settings where tasks are imbalanced and multi-labelled. Compared to state-of-the-art models, DynaGraph achieves significant improvements in balanced accuracy and sensitivity over the nearest complex competitors in time-series or dynamic graph modelling across three ICU and one primary care datasets. Through a pseudo-attention approach to graph construction, our model also indicates the importance of clinical covariates over time, providing means for clinical validation.
\end{abstract}

\section{INTRODUCTION}
The increasing availability of electronic health records (EHRs) and time-series data offers unprecedented opportunities to assist clinicians in diagnosing diseases and predicting patient outcomes, ultimately improving treatment plans \citep{siontis2021artificial, smith2014early, shamout2020machine}. However, robustly modelling EHRs requires addressing significant challenges, including noisy, sparse, irregularly sampled, and complex multivariate time-series data, while simultaneously predicting multiple outcomes. Beyond handling these complexities, interpretability is critical, as clinician trust in model predictions is essential to influence treatment decisions \citep{lauritsen2020explainable}.

Traditional classification methods rely on manual feature extraction, which is limited by domain expertise and time constraints. Deep learning approaches, such as long- and short-term memory (LSTM) networks, gated recurrent units (GRUs), transformers, and temporal convolutional networks, have been widely adopted to automatically extract temporal patterns from EHRs for tasks such as risk prediction, phenotyping, and clustering \citep{duffy2022high, lee2020temporal, aguiar2022learning, rafiei2021ssp, tan2020data, li2020behrt, kok2020automated}. However, these methods often fail to explicitly capture hidden dependencies between multivariate time-series or model the intricate spatial and temporal interdependencies \citep{duan2022multivariate, fan2025medgnn}.

Graph representation learning offers a promising solution by modelling these dependencies through dynamic graph structures. Recent approaches in temporal graph learning, such as sequence-based models and graph neural networks (GNNs), often rely on predefined graph structures, limiting their ability to capture evolving relationships in EHR data \citep{wang2022sparsification, bogaerts2020graph, rocheteau2021predicting, zhou2020graph, huang2023benchtemp}. Spatio-temporal GNNs, which integrate spatial and temporal information, have shown potential but are constrained by their reliance on static or manually constructed graphs \citep{han2021dynamic, liu2023todynet}. We propose constructing dynamic, adaptive graph structures that learn spatio-temporal representations through information propagation and node/edge completion to address these limitations. Our approach aggregates temporal embeddings from sequential modules with learnable graph matrices, capturing feature correlations over time. Additionally, we introduce a pseudo-attention mechanism during graph construction to focus on the most important temporal patterns, enabling the model to learn meaningful representations without assuming a predefined graph structure.

In this work, we present DynaGraph, a contrastive, interpretable dynamic graph model designed for challenging multi-label imbalanced prediction settings for multivariate EHR time-series data. DynaGraph advances the state-of-the-art by:

\begin{itemize}
\item Introducing a novel spatio-temporal graph approach that combines sequential embeddings, information propagation, and pseudo-attention. Our model dynamically constructs graphs without predefined structure or constraints to capture hidden dependencies in multivariate time-series.
\item Combining interpretable dynamic graph learning, temporal recurrence, and variational graph learning into a unified end-to-end framework for joint optimisation, ensuring robust predictive performance.
\item Proposing an intuitive interpretability mechanism for learning dynamic graph representation, enabling time-resolved feature importance analysis.
\item Exploring a novel graph augmentation strategy coupled with contrastive loss to enhance prediction performance.
\item Combining focal and structural losses to stabilise dynamic graph learning in time-series imbalanced multi-label settings.
\end{itemize}

We evaluated DynaGraph in four real-world healthcare datasets from the secondary ICU and primary care settings, demonstrating superior performance over state-of-the-art methods in broad EHR time-series tasks, as well as dynamic graph models. Our results highlight the model's potential for clinical applications, particularly in settings requiring interpretability and robust performance on imbalanced tasks.

\section{RELATED WORK}

\subsection{Multivariate EHR Time-Series Classification}
Multivariate time-series classification in electronic health records (EHRs) presents unique challenges due to the high dimensionality, irregular sampling, and inherent missingness of the data. Traditional approaches often rely on recurrent neural networks (RNNs), such as LSTMs and GRUs, which have been widely applied to EHR tasks \citep{harutyunyan2019multitask, sheikhalishahi2019benchmarking}. Transformer-based architectures have emerged as powerful alternatives, demonstrating improved performance in capturing long-range dependencies and complex temporal patterns in EHR datasets \citep{wang2022integrating, luo2023pt3}. Specialised models, such as T-LSTM, RETAIN, and BiT-MAC, have also been proposed, extending these architectures to better handle the unique characteristics of the EHR data \citep{baytas2017patient, choi2016retain, wang2023bit}. In particular, Medformer introduced a patching transformer designed to capture intricate temporal dynamics in multivariate time-series \citep{wang2024medformer}. However, these methods often fail to explicitly model inter-feature dependencies or address the pervasive issue of missing data. Furthermore, extending these approaches to multi-label classification, where patients may experience multiple concurrent outcomes, remains a significant challenge, as most models are designed for single-label tasks.

\subsection{Graph Neural Networks}
Graph Neural Networks (GNNs) offer a promising framework to address these limitations by explicitly modelling structural dependencies between time-series features, even in the presence of missing data \citep{wu2020connecting, bai2020adaptive}. In GNNs, node representations are transformed and aggregated with their neighbours, allowing the capture of complex dependencies across features and time steps \citep{xu2021graph}. By representing each time-series as a node in a graph, GNNs can effectively model spatial-temporal relationships. Popular variants, such as Graph Convolutional Networks (GCNs), Graph Attention Networks (GATs), and GraphSAGE, introduce different mechanisms for node transformation and neighbourhood aggregation \citep{kipf2016semi, velivckovic2017graph, hamilton2017inductive}. However, these methods typically require a predefined graph structure, which is often unavailable in real-world scenarios, limiting their applicability. Moreover, these approaches lack interpretability and are constrained by their reliance on static graph structures. Here, interpretability refers to the ability to establish clear associations between time-series features and model predictions.

Spatio-temporal GNNs extend this paradigm by capturing spatial dependencies through graph convolutions and temporal dependencies via recurrent neural networks or temporal convolutions. However, existing models, such as STFGNN and TAGNet, are limited by their reliance on predefined graph structures, which can hinder their ability to capture intrinsic data dynamics \citep{li2021spatial, wang2020tagnet}. SimTSC addresses some of these limitations by leveraging Dynamic Time Warping to model pairwise feature similarities, improving time-series classification through similarity patterns across time points and features \citep{zha2022towards}. However, it still relies on predefined graph structures and DTW constraints. TodyNet, while addressing some of these constraints, suffers from computational inefficiency, limited temporal embedding capabilities, and lack of interpretability \citep{liu2023todynet}. MedGNN proposes a multi-resolution approach to non-EHR multichannel time-series such as EEGs and ECGs, where the graphs are pre-defined by embeddings from channel-wise convolutions, similar to TodyNet, and do not address the interpretability challenge \citep{fan2025medgnn}.

In this work, we propose a novel spatio-temporal graph modelling framework that dynamically constructs graph representations directly from complex, heterogeneous multivariate time-series in an end-to-end manner. Our approach addresses the limitations of predefined graph structures, enhances interpretability, and adapts to challenging multi-label imbalanced classification scenarios, offering a significant advancement over existing methods in the literature.

\section{Data}
We evaluated our proposed framework on four publicly available longitudinal electronic health record (EHR) datasets that include both ICU and primary care settings. The MIMIC-III v1.4 ICU dataset comprises 53,423 distinct hospital admissions \citep{johnson2016mimic}. Following the preprocessing pipeline proposed by \citep{wang2020mimic}, we extract a subcohort of 17,279 patients, each characterised by static variables (age and sex) and 42 laboratory measurements in time-series. The prediction tasks include ten binary classification outcomes: ICU mortality (7.65\% positive cases), hospital mortality (9.65\%), 30-day readmission (2.19\%), shock (7.57\%), acute cerebrovascular disease (8.99\%), acute myocardial infarction (10.26\%), cardiac dysrhythmias (32.60\%), chronic kidney disease (12.14\%), chronic obstructive pulmonary disease and bronchiectasis (11.82\%), and congestive heart failure (24.15\%). The eICU Collaborative Research Database is a multi-centre ICU dataset containing over 200,859 patient unit encounters from 139,367 unique patients across 335 ICUs in 208 hospitals in the United States \citep{pollard2018eicu}. We focus on a subset of heart attack patients admitted to critical care units, such as the coronary care unit (CCU), who were monitored for potential complications. Among these patients, 16.00\% developed at least one of the following complications: peripheral vascular disease (1.10\%), heart failure (10.26\%), atrial fibrillation (28.49\%), arrhythmias (23.14\%), or death (12.00\%) \citep{elbadawi2019temporal}. The HiRID-ICU dataset includes more than 33,000 patient admissions to an intensive care unit. We used the imputed staging dataset provided by the original investigators, with prediction tasks that focused on ICU mortality (8.39\%), respiratory failure (38.12\%), and circulatory failure (3.10\%) \citep{hyland2020early, yeche2021hirid}. EHRSHOT is based on primary care data from the Stanford Medicine Research Data Repository and Stanford Health Care of 6,739 patients \citep{wornow2024ehrshot}. After removing patients with fewer than 24 time-stamped measurements, no time-series vital signs or laboratory values, outliers, and no missing labels of interest, 2,378 remain. Our labels include the first occurrence of heart attack (9.52\%), lupus (5.73\%), celiac disease (3.45\%), pancreatic cancer (9.30\%), hyperlipidemia (14.84\%), and hypertension (16.39\%) after 1 year post-discharge.

For all datasets, we divide the data into training, validation, and test sets based on patient IDs, using an 8: 1: 1 ratio. Additional details on preprocessing are provided in Supplementary Section 1.

\section{METHODS}

\begin{figure*}[t]
    \centering
    \includegraphics[width=1.0\linewidth]{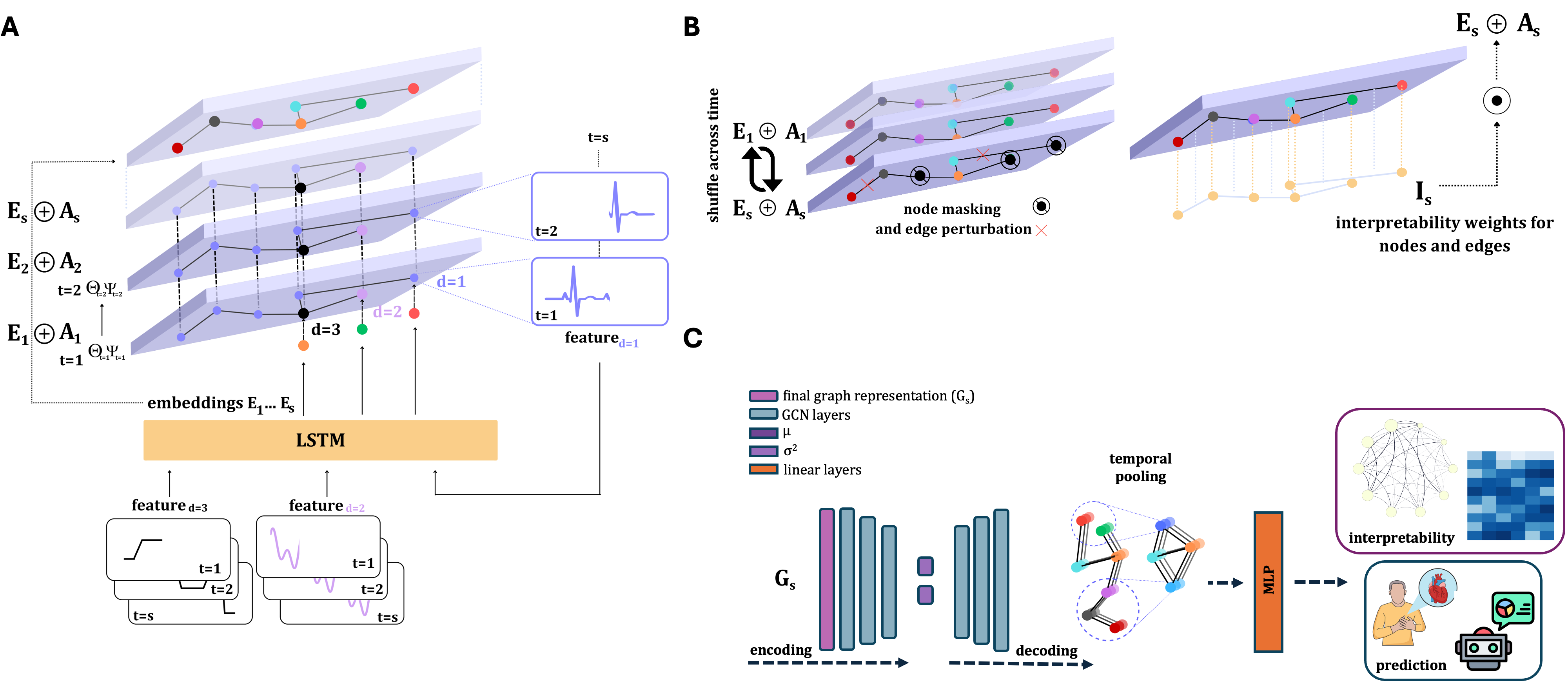}
    \caption{Our DynaGraph model framework. The multivariate time-series $x_1, x_2, \ldots, x_d$ are divided into \textit{s} equal-length time windows $t_1, t_2, \ldots, t_s$. \textbf{A:} Each window $t_1, t_2, \ldots, t_s$ has a corresponding feature matrix represented by a dynamic graph $A_1, A_2, \ldots, A_s$, whose adjacency matrices are learnable through information propagation. The node and edge vectors of the previous graph slice, $\theta_{t-1}, \psi_{t-1}$, respectively, inform the corresponding nodes and edges of the next graph slice $\theta_{t}, \psi_{t}$. The connection (expressed as dotted lines) from node to node between graph slices corresponds to information propagation across time. An embedding matrix from the LSTM output $E_1, E_2, ..., E_s$ for the same time windows as for the graph construction ensures the pairing between the graphs and the temporal embeddings. \textbf{B:} The adjacency matrix is paired with an interpretability weight matrix $I$ whose weights are learnable for every slice with a Hadamard product. The interpretability weights measure the impact of the different parts of the graph on the final loss $\mathcal{L}_{\text {total}}$. The adjacency matrix is also augmented with graph augmentation techniques for contrastive loss computation, such as shuffling across the $s$ time dimension, random node masking, and edge perturbations. The adjacency matrix, interpretability matrix, and embeddings are aggregated to produce the final spatio-temporal graph representation of the multivariate time-series, namely $G_1, G_2, \ldots, G_s$. \textbf{C:} Finally, $G_s$ is passed through a VGAE with GINs as the encoder-decoder and its output graph representation is clustered temporally with CNNs before being flattened for a standard MLP for multi-label classification.}
    \label{fig:DynaGraph}
\end{figure*}

\subsection{Dynamic Graph Construction}
Figure \ref{fig:DynaGraph} shows the overall framework of DynaGraph. Input is defined as a collection of multivariate time-series measurements for $N$ patients. A patients is then described by $X=\left\{x_1, x_2, \ldots, x_d\right\} \in$ $\mathbb{R}^{d \times l}$ with $d$ features of length $l$. Given a group of $m$ patients $X=\left\{X_1, X_2, \ldots, X_m\right\} \in \mathbb{R}^{m \times d \times l}$, their corresponding labels $Y=\left\{y_1, y_2, \ldots, y_m\right\}$ mean $y$ is a predefined class vector of binary labels for each patient, and $m \in \mathbb{N}^*$. The final graph construction consists of three components:

\paragraph{Graph Construction Through Information Propagation.} We dissect the time-series into \textit{s} equal-sized time-windows  $T=\left[t_1, t_2, \ldots, t_s\right]$, $X$ now becomes $X \in$ $\mathbb{R}^{d \times l \times s}$, where the first window $t_1$ is used to construct an initial static graph representation with nodes representing the time-series features and the edges the hidden associations between the features. The graph representation is captured in an adjacency matrix with the rows and columns corresponding to the features and nodes, respectively. All elements of the adjacency matrix are learnable parameters in the model, initialised randomly. Each node is assigned two values, the source and target nodes. We generate vectors $\Theta$ and $\Psi$ with length $d$ (number of features) for each time window $t$, and all elements are learnable parameters that are initialised randomly. The initial adjacency matrix is then the multiplication of these vectors for a time slot:
\begin{equation}
A=\Theta^{\mathrm{T}} \cdot \Psi \in \mathbb{R}^{d \times d}
\end{equation}
where $\Theta=\left[\theta_{t, 1}, \theta_{t, 2}, \ldots, \theta_{t, d}\right], \Psi=\left[\psi_{t, 1}, \psi_{t, 2}, \ldots, \psi_{t, d}\right]$ represent the random initialization of learnable node embeddings. The adjacency matrix is made more sparse to reduce the computational costs by using the top-k largest values of the adjacency matrix:
\begin{equation}
\begin{gathered}
i d x, i d y=\operatorname{argtopk}(A[:,,]) \quad i d x \neq i d y \\
A[-i d x,-i d y]=0
\end{gathered}
\end{equation}
Subsequent time windows are used to construct dynamic graph representations, aggregated node-wise through message passing. For each time slot, new vertices are added to represent features from the previous time slot, resulting in a set of vertices $\left\{v_{(t, 1)}, v_{(t, 2)}, \ldots, v_{(t, D)}, v_{(t-1,1)}, v_{(t-1,2)}, \ldots, v_{(t-1, D)}\right\}$. The edges are directed from previous time vertices to their counterparts in the current time window, connecting $v_{(t-1, d)}$ to $v_{(t, d)}$ for $d=$ $1,2, \ldots, D$ where $D$ is the total number of features or nodes. Since the number of features, nodes, or vertices does not change over time, the new set of nodes is connected to the previous set, signifying the addition of time connections in the graph. To prevent an exponential increase in the number of nodes, new node embeddings are aggregated, and redundant vertices are removed. The graph representation is a set of adjacency matrices $A=\left\{A_1, A_2, \ldots, A_s\right\} \in$ $\mathbb{R}^{d \times d \times s}$ with $d$ features or nodes and for $s$ time-windows capturing spatio-temporal patterns in the multivariate time-series.

\paragraph{LSTM Embeddings.} $X \in \mathbb{R}^{d \times l \times s}$ is also, in parallel, processed by a Long-Short Term Memory (LSTM) unit to derive an embedding matrix for each time slice, encapsulated as $E = \{E_1, E_2, ..., E_s\} \in \mathbb{R}^{d \times d \times s}$. The LSTM embeddings help capture the long-term intra-feature temporal patterns of individual time-series whereas the graph learns efficient inter-feature correlations in discrete time slices.  

\paragraph{Interpretability.} We used a paired weight matrix for each graph with learnable weights that would update itself based on the contribution of each node or edge of the adjacency matrix to the total loss. The adjacency matrix of each graph is paired with a uniformly initialised weight matrix $I$ of the same dimension $d \times d \times s$ whose weights are updated depending on which node or edge contributes the most to the downstream loss computation. This allows us to use the weights in this matrix as 'pseudo-attention' weights, telling us which parts of the adjacency matrix, and thus the graph, contributed the most to the prediction. A visualisation of this can be seen in Supplementary Figure 1 where the learnable weights are paired with the node columns or features. 

We define a set of interpretability weight matrices $I=\left\{I_1, I_2, \ldots, I_s\right\} \in$ $\mathbb{R}^{d \times d \times s}$ on top of the adjacency matrices for each time slice whose weights are updated based on the graph nodes or edges contributing to the loss gradient update. More details are described in Supplementary Section \textit{Graph Isomorphism Network (GIN) and Interpretability}. 

The total contribution of a feature (node) \( v \), $I_v$, is quantified by combining its direct importance with the average importance of its connections (edges) to other nodes. This can be expressed as:
\begin{equation}
    I_v = \alpha \cdot I_{vv} + (1-\alpha) \cdot \bar{I}_{vu}
\end{equation}
where \( \alpha \) is a balancing parameter. The node and edge importances are elements in the interpretability matrix and are computed as follows:
\begin{equation}
  I_{vv} = \| \nabla_{h_v} L \|
\end{equation}
\begin{equation}
   I_{vu} = \| \nabla_{e_{v,u}} L \|
\end{equation}
where \( h_v \) denotes the feature vector of node \( v \), \( e_{v,u} \) represents the edge between nodes \( v \) and \( u \), and \( L \) is the loss function of the model. \( \text{Average Edge Importance}_v \) is the average of the gradients for the edges connected to node \( v \). After iteration, the final values of the importance weights are normalised across all features to ensure comparability. We show how the interpretability weight matrix $I$ converges under certain loss and learning rate assumptions in Supplementary Section \textit{Graph Isomorphism Network (GIN) and Interpretability}.

The adjacency, embedding, and interpretability matrices are aggregated for a final graph representation $G$:
\begin{equation}
    G^{(i)} = \left( A^{(i)} \parallel I^{(i)} \parallel E^{(i)} \right)
\end{equation}
The embeddings are added to the adjacency matrices, and the interpretability weights are multiplied with a Hadamard product, respectively. This final representation then captures both the temporal patterns within individual time-series from the LSTM, the spatio-temporal patterns between time-series features from the graph construction, and the time-feature importance. Details are described further in Supplementary Sections 3 and 5.

\subsection{Graph Augmentation, Model Training and Interpretation}
We introduce graph augmentations to help our model generalise and learn temporal variability. We couple this with a contrastive loss which encourages the model to capture both spatial (inter-node) and temporal (intra-node) dependencies. This approach is described in Supplementary Figure 1 and augmentations include shuffling along the time axis and perturbing the adjacency matrix $A$ through node dropping or edge perturbations, creating an augmented graph pair $A_{+}$. A negative sample $A_{-}$ is also created for each original graph pair $A$. The augmented and negative samples are compared to the original graph using cosine similarity, forming the contrastive loss:
\begin{equation}
\mathcal{L}_{\text {contrast }}=-\mathbb{E}\left[\log \frac{e^{\operatorname{Sim}_\theta\left(A, A_{+}\right)}}{e^{\operatorname{Sim}_\theta\left(A, A_{+}\right)}+\sum_{A_{-}} e^{\operatorname{Sim}_\theta\left(A, A_{-}\right)}}\right]
\end{equation}
$\operatorname{Sim}_\theta$ represents the cosine similarity $\operatorname{Sim}_\theta\left(\boldsymbol{x}_{n, i}, \boldsymbol{x}_{n, j}\right)=\boldsymbol{x}_{n, i}^{\top} \boldsymbol{x}_{n, j} /\left\|\boldsymbol{x}_{n, i}\right\|\left\|\boldsymbol{x}_{n, j}\right\|$ for the pair of nodes $i,j$ of the $n$th graph or adjacency matrix in the minibatch. 

To address the class imbalance in multi-label prediction, we employ a focal loss:
\begin{equation}
\mathcal{L}_{\text {focal }}(\widehat{y})=-(1-\widehat{y})^\gamma \cdot \log (\widehat{y}), \quad \gamma \geq 0
\end{equation}
where the $\gamma$ parameter accounting for the class weighting is considered as a hyperparameter.

A regularization term penalizes large variations in node features to preserve smoothness:
\begin{equation}
\mathcal{L}_{\text {reg }} = \lambda \sum_{(i, j) \in E} \| \mathbf{h}_i - \mathbf{h}_j \|^2
\end{equation}
where $\lambda$ is a hyperparameter controlling the strength of the regularization, $(i, j)$ represents an edge connecting nodes $i$ and $j$, and $\mathbf{h}_i$, $\mathbf{h}_j$ are the feature representations of nodes $i$ and $j$, respectively.

Additionally, we define a structural similarity loss to ensure that the learned graph structure remains close to the original (previous timepoint) structure:
\begin{equation}
\mathcal{L}_{\text {structure }} = \mu \left(1 - \frac{\sum_{i,j} A_{ij} \cdot A'_{ij}}{\sqrt{\sum_{i,j} A_{ij}^2} \cdot \sqrt{\sum_{i,j} A_{ij}^{\prime2}}}\right)
\end{equation}
where $A$ is the original adjacency matrix, $A'$ is the adjacency matrix after augmentation, and $\mu$ is a hyperparameter. The structural loss aids in the convergence of the adjacency matrix throughout learning. The final loss is the sum of these losses where the balancing parameters $\epsilon$, $\lambda$, $\mu$, and $\beta$ are considered as hyperparameters:

\begin{equation}
\begin{gathered}
\mathcal{L}_{\text {total }} =\mathcal{L}_{\text {BCE }} + \alpha \mathcal{L}_{\text {contrast }} + \epsilon \mathcal{L}_{\text {focal }} + \lambda \mathcal{L}_{\text {reg }} + \\ \mu \mathcal{L}_{\text {structure }} + \beta \mathcal{L}_{\text{VGAE}}
\end{gathered}
\end{equation}

As described earlier in this section, focal loss is a common adjustment made in cases of strong class imbalance as in our healthcare scenarios. Contrastive loss aids in predictive performance by allowing for more diverse or augmented representations of the input graphs in learning downstream tasks. The regularisation loss term is added to reduce overfitting and stabilise the loss within bounds. The structural loss ensures that next time-step graph representations do not deviate too far from the previous representation to stabilise the dynamic graphs through time.

After integrating the embeddings with the augmentations, $G_{s}$ is passed through a variational graph autoencoder (VGAE) whose encoder-decoder structure follows a GIN architecture. The latent features are graph representations of the mean and variance vectors, and further details, as well as on the loss formulation due to space constraints, can be found in Supplementary Section \textit{VGAE}. The output of the VGAE is a graph reconstruction which is pooled temporally to reduce the number of nodes with dynamic clustering and decrease the computational costs of the training. The reduced graph is then flattened and passed through a multilayer perceptron for the final multilabel classification.

\section{RESULTS}

\subsection{Models Comparision}
We compare DynaGraph with the state-of-the-art models for time series classification in healthcare. We looked at two main groups of models drawing from related recent work: time-series models and dynamic graph models. LSTMs, Gates Recurrent Units (GRUs), Temporal Convolution Networks (TCNs), Transformers, T-LSTM, RETAIN, and BiT-MAC were from the first group. The graph models we used were Graph Convolutional Networks (GCNs), Graph Attention Networks (GATs), TodyNet, SimTSC, and MedGNN. We compare these models using metrics of balanced accuracy (due to high-class imbalance present in these tasks and cohorts), F1-score (commonly used in multi-label settings), and sensitivity to detect the propensity of these models to identify high-risk patients. The details of their implementations can be found in Supplementary Section 1. All experiments are implemented using PyTorch 3.8 and are performed on a NVIDIA A100 Tensor Core GPU. The repository to implement the model can be found here: \url{https://github.com/munibmesinovic/DynaGraph.git}. Table \ref{tab:Result} presents a comparison of DynaGraph with benchmark models across four datasets. As shown, DynaGraph consistently outperforms all other models across all datasets and various care settings, with the most notable improvement observed in sensitivity. Using contrastive and focal losses along with graph augmentations, DynaGraph enhances the detection of positive cases without compromising specificity, as reflected in a comparable F1 score. Furthermore, its superior balanced accuracy in highly imbalanced scenarios highlights not only its accuracy but also its robustness in handling imbalanced healthcare tasks.

\begin{table*}[!]
    \small
    \centering
    \caption{Test results on different datasets for time-series and graph models in multi-label classification. For accuracy, we used balanced accuracy due to severe class imbalance. Sens stands for sensitivity averaged across all labels. For all metrics, the higher the better. The best results are in bold. The results are an average of five random seeds.}
    \label{tab:Result}
    \setlength{\tabcolsep}{5pt}
    \arrayrulecolor{gray!40}  

    \begin{subtable}[t]{0.45\textwidth}
        \centering
        \caption{MIMIC-III}
        \begin{tabular}{lccc}
            \toprule
            Model & Acc* & F1 & Sens \\
            \midrule
            LSTM       & 65.18 $\pm 0.31$ & 38.46 $\pm 0.20$ & 47.12 $\pm 0.69$ \\
            \midrule
            GRU        & 68.38 $\pm 0.53$ & 47.55 $\pm 0.47$ & 42.70 $\pm 0.74$ \\
            \midrule
            Transformer & 69.16 $\pm 0.54$ & 44.63 $\pm 0.46$ & 48.81 $\pm 0.79$ \\
            \midrule
            TCN        & 70.94 $\pm 0.43$ & 45.88 $\pm 0.47$ & 59.03 $\pm 0.88$\\
            \midrule
            T-LSTM	&67.30 $\pm 0.48$	&40.50 $\pm 0.58$	&48.92 $\pm 0.85$ \\
            \midrule
            RETAIN &70.67 $\pm 1.37$	&45.97 $\pm 1.45$	&66.30 $\pm 2.39$ \\
            \midrule
            BiT-MAC	&68.32 $\pm 1.22$	&43.60 $\pm 1.30$	&47.25 $\pm 2.06$ \\
            \midrule
            GAT        & 66.47 $\pm 0.51$ & 41.79 $\pm 0.59$ & 47.16 $\pm 0.99$ \\
            \midrule
            GCN        & 69.30 $\pm 0.54$ & 42.83 $\pm 0.66$ & 51.24 $\pm 1.03$ \\
            \midrule
            TodyNet        & 71.80 $\pm 0.32$ & 44.67 $\pm 0.25$ & 59.91 $\pm 0.73$ \\
            \midrule
            SimTSC        & 70.91 $\pm 0.78$ & 42.38 $\pm 0.42$  & 57.75 $\pm 1.15$ \\
            \midrule
            MedGNN        & 77.63 $\pm 0.37$ & 46.08 $\pm 0.28$ & 73.20 $\pm 0.94$ \\
            \midrule
            \rowcolor{gray!20} DynaGraph & \textbf{79.29 $\pm 0.31$} & \textbf{47.04 $\pm 0.23$} & \textbf{85.22 $\pm 0.76$} \\
            \bottomrule
        \end{tabular}
    \end{subtable}%
    \hfill
    \begin{subtable}[t]{0.45\textwidth}
        \centering
        \caption{eICU}
        \begin{tabular}{lccc}
            \toprule
            Model & Acc* & F1 & Sens \\
            \midrule
            LSTM       & 60.33 $\pm 0.36$& 30.07 $\pm 0.19$ & 39.58 $\pm 0.59$ \\
            \midrule
            GRU        & 58.43 $\pm 0.63$ & 29.91 $\pm 0.52$ & 30.09 $\pm 0.79$\\
            \midrule
            Transformer & 61.72 $\pm 0.52$ & 32.74 $\pm 0.48$ & 35.49 $\pm 0.78$\\
            \midrule
            TCN        & 64.12 $\pm 0.47$ & 38.61 $\pm 0.40$ & 40.83 $\pm 0.77$\\
            \midrule
            T-LSTM	&63.90 $\pm 0.46$	&32.48 $\pm 0.51$	&40.76 $\pm 0.78$\\
            \midrule
            RETAIN	&63.72 $\pm 1.40$	&36.00 $\pm 1.48$	&40.10 $\pm 2.57$\\
            \midrule
            BiT-MAC	&62.40 $\pm 1.07$	&35.55 $\pm 1.18$	&39.44 $\pm 1.84$ \\
            \midrule
            GAT        & 66.10 $\pm 0.47$ & 31.84 $\pm 0.53$ & 36.74 $\pm 0.82$ \\
            \midrule
            GCN        & 68.80 $\pm 0.49$ & 32.96 $\pm 0.60$ & 37.91 $\pm 0.93$ \\
            \midrule
            TodyNet        & 69.92 $\pm 0.32$ & 40.25 $\pm 0.26$ & 78.11 $\pm 0.75$\\
            \midrule
            SimTSC        & 69.33 $\pm 0.72$ & 39.80 $\pm 0.44$ & 70.29 $\pm 1.31$ \\
            \midrule
            MedGNN        & 73.11 $\pm 0.36$ & 44.67 $\pm 0.37$ & 81.22 $\pm 0.99$ \\
            \midrule
            \rowcolor{gray!20} DynaGraph & \textbf{73.52$\pm 0.29$} & \textbf{45.96 $\pm 0.23$} & \textbf{86.00$\pm 0.65$} \\
            \bottomrule
        \end{tabular}
    \end{subtable}

    \vspace{10pt}

    \begin{subtable}[t]{0.45\textwidth}
        \centering
        \caption{HiRID-ICU}
        \begin{tabular}{lccc}
            \toprule
            Model & Acc* & F1 & Sens \\
            \midrule
            LSTM       & 78.22$\pm 0.46$ & 42.48$\pm 0.29$ & 67.10$\pm 0.70$ \\
            \midrule
            GRU        & 77.30$\pm 0.59$ & 44.11$\pm 0.49$ & 66.60$\pm 0.90$ \\
            \midrule
            Transformer & 77.57$\pm 0.56$ & 46.47$\pm 0.63$ & 65.20$\pm 0.80$ \\
            \midrule
            TCN        & 81.98$\pm 0.52$ & 52.29$\pm 0.57$ & 70.80$\pm 0.80$ \\
            \midrule
            T-LSTM	&78.81$\pm 0.54$	&43.27$\pm 0.56$	&69.00$\pm 0.97$ \\
            \midrule
            RETAIN	&80.33$\pm 1.05$	&51.89$\pm 1.11$	&69.27$\pm 1.93$ \\
            \midrule
            BiT-MAC	&78.90 $\pm 1.28$	&49.31$\pm 1.30$	&67.10$\pm 1.92$ \\
            \midrule
            GAT        & 73.48 $\pm 0.47$ & 40.95 $\pm 0.53$ & 61.30 $\pm 0.82$ \\
            \midrule
            GCN  & 74.36$\pm 0.55$ & 41.80$\pm 0.61$ & 66.50$\pm 0.97$ \\
            \midrule
            TodyNet & 79.91$\pm 0.45$ & 57.13$\pm 0.40$ & 78.22 $\pm 0.89$\\
            \midrule
            SimTSC        & 77.75$\pm 0.84$ & 54.12$\pm 0.51$  & 71.90 $\pm 1.20$\\
            \midrule
            MedGNN        & 81.29$\pm 0.39$ & 52.57$\pm 0.41$ & 82.55 $\pm 1.07$\\
            \midrule
            \rowcolor{gray!20} DynaGraph & \textbf{84.30$\pm 0.34$} & \textbf{59.37$\pm 0.27$} & \textbf{86.20$\pm 0.83$} \\
            \bottomrule
        \end{tabular}
    \end{subtable}%
    \hfill
    \begin{subtable}[t]{0.45\textwidth}
    \centering
    \caption{EHRSHOT}
    \begin{tabular}{lccc}
        \toprule
        Model & Acc* & F1 & Sens \\
        \midrule
        LSTM       & 56.70$\pm 0.39$ & 31.91$\pm 0.30$ & 53.40$\pm 0.70$ \\
        \midrule
        GRU        & 57.51$\pm 0.50$ & 32.24$\pm 0.47$ & 54.73$\pm 0.76$ \\
        \midrule
        Transformer & 56.29$\pm 0.59$ & 30.88$\pm 0.49$ & 52.90$\pm 0.80$ \\
        \midrule
        TCN        & 61.27$\pm 0.48$ & 37.30$\pm 0.45$ & 60.11$\pm 0.85$ \\
        \midrule
        T-LSTM	&65.17$\pm 0.49$	&40.08$\pm 0.55$	&64.82$\pm 0.86$ \\
        \midrule
        RETAIN	&69.48$\pm 1.39$	&42.33$\pm 1.47$	&67.96$\pm 2.38$ \\
        \midrule
        BiT-MAC	&68.36$\pm 1.23$	&41.23$\pm 1.28$	&66.14$\pm 2.07$ \\
        \midrule
        GAT        & 64.55 $\pm 0.50$ & 37.82 $\pm 0.58$ & 62.39 $\pm 0.98$ \\
            \midrule
        GCN  & 64.79$\pm 0.55$ & 39.74$\pm 0.67$ & 63.21$\pm 1.01$ \\
        \midrule
        TodyNet & 69.25$\pm 0.33$  & 41.98$\pm 0.26$ & 66.82$\pm 0.76$ \\
        \midrule
        SimTSC        & 71.41$\pm 0.77$ & 42.96$\pm 0.41$ & 69.37$\pm 1.14$ \\
        \midrule
        MedGNN        & 77.46$\pm 0.36$ & 44.91$\pm 0.27$ & 74.23$\pm 0.95$ \\
        \midrule
        \rowcolor{gray!20} DynaGraph & \textbf{79.83$\pm 0.32$} & \textbf{46.10$\pm 0.24$} & \textbf{79.44$\pm 0.71$} \\
        \bottomrule
    \end{tabular}
\end{subtable}

\end{table*}

\subsection{Model Interpretability}
\begin{figure*}
        \centering
        \begin{subfigure}[b]{0.32\textwidth}   
            \centering 
            \includegraphics[width=\textwidth]{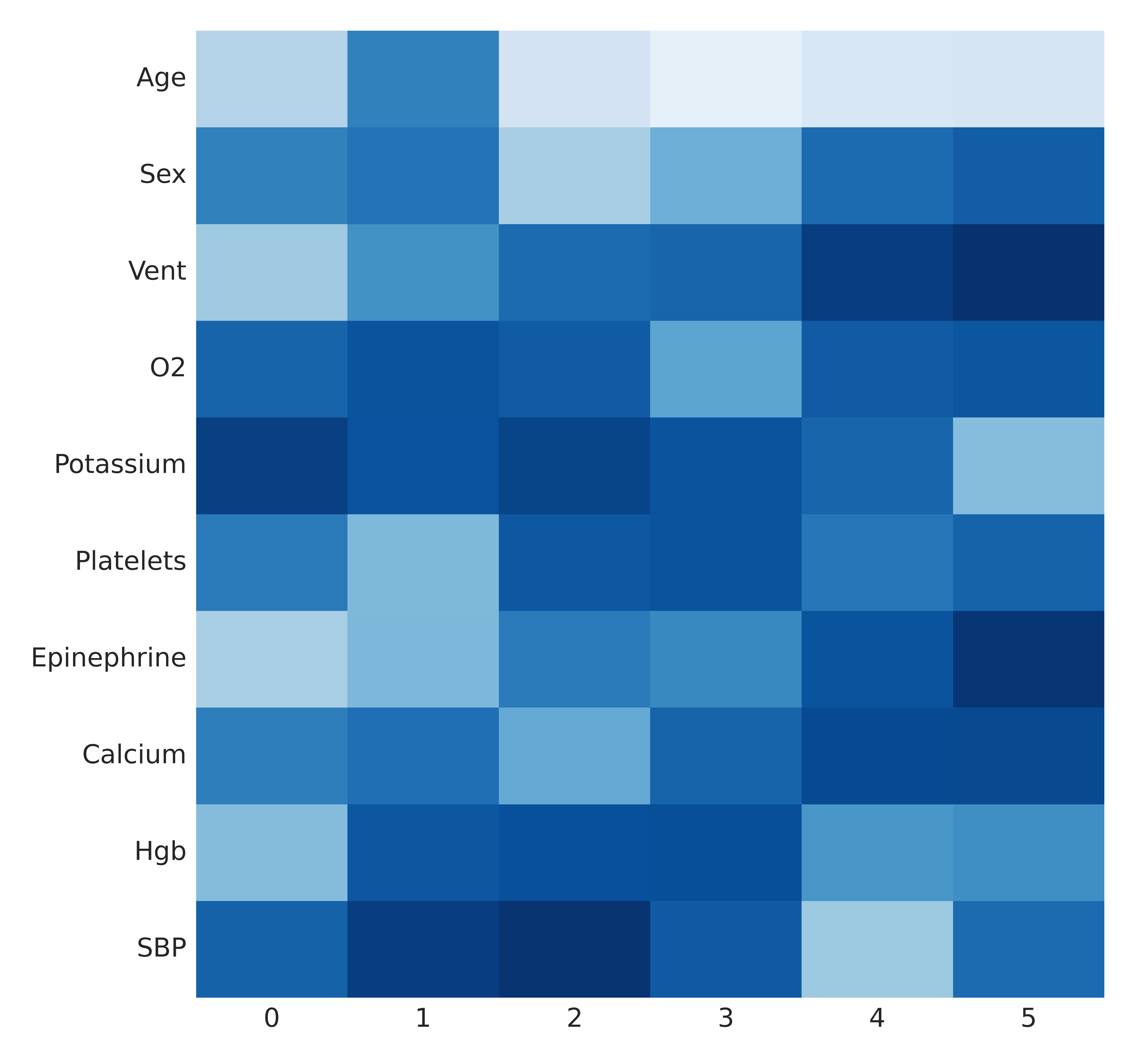}
            \caption[]%
            {{\small MIMIC-III}}    
        \end{subfigure}
        \hfill
        \begin{subfigure}[b]{0.32\textwidth}   
            \centering 
            \includegraphics[width=\textwidth]{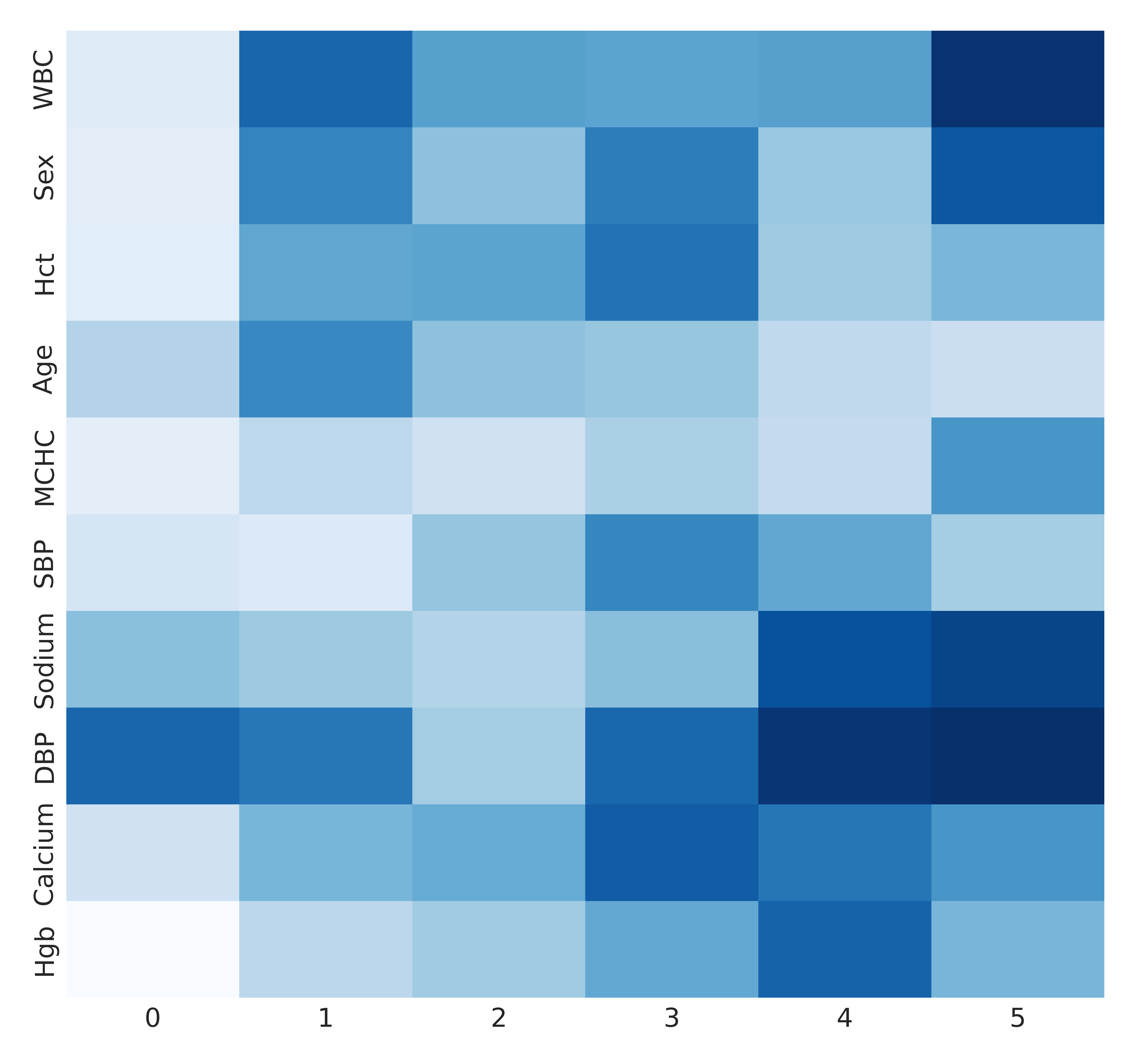}
            \caption[]%
            {{\small eICU}}
            \label{fig:mean and std of net44}
        \end{subfigure}
        \hfill
        \begin{subfigure}[b]{0.32\textwidth}   
            \centering 
            \includegraphics[width=\textwidth]{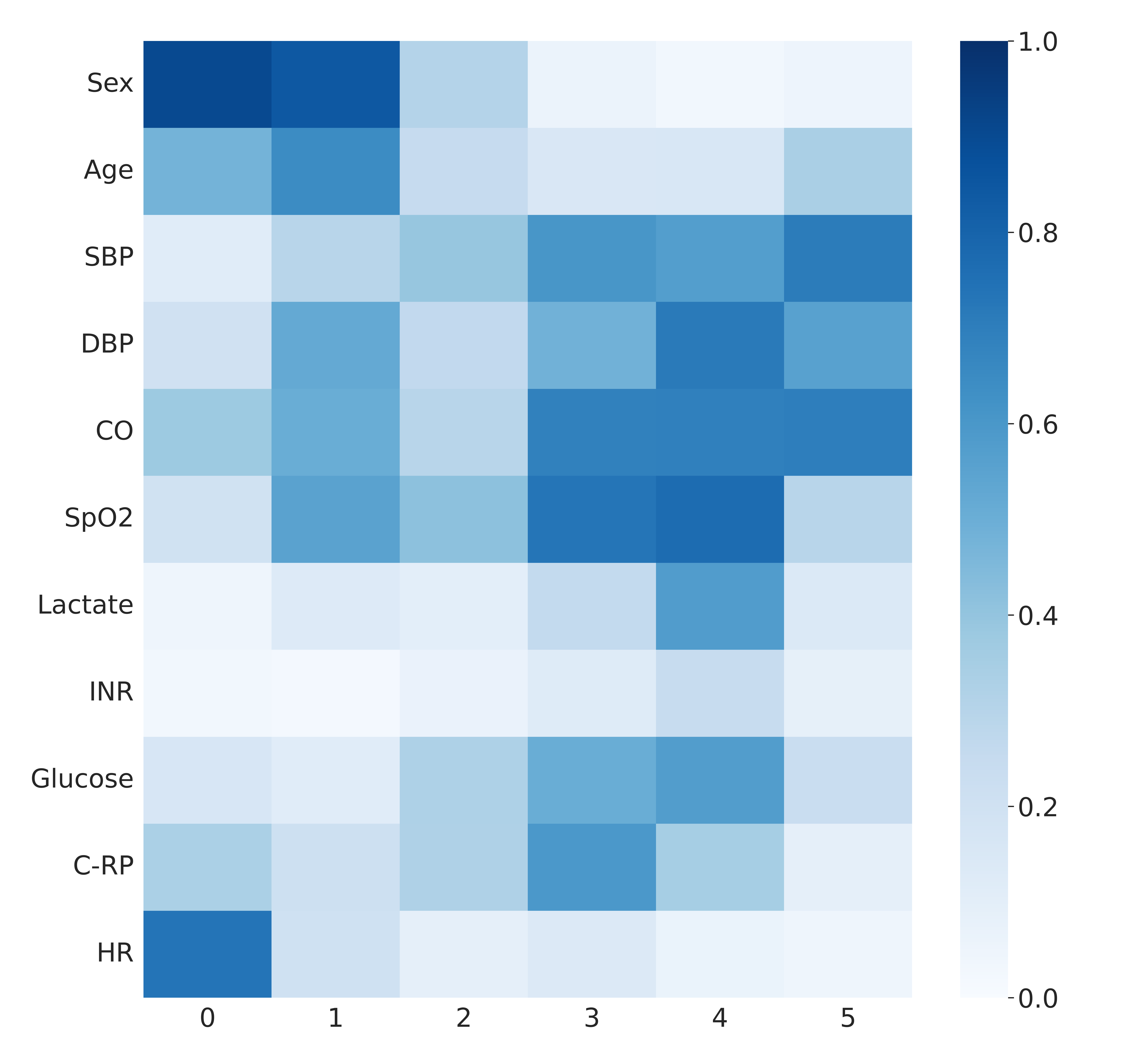}
            \caption[]%
            {{\small HiRID-ICU}}
            \label{fig:mean and std of net44}
        \end{subfigure}
        \caption
        {\small Heatmaps of the pseudo-attention weight matrices for DynaGraph during training on (a) MIMIC-III, (b) eICU, and (c) HiRID datasets, highlighting the globally top 10 features. The weights are normalized and smoothed using a Gaussian kernel ($\sigma=0.6$). Higher values indicate greater feature importance for the corresponding time-period in the final multi-label prediction tasks: heart attack complications (eICU), phenotype classification (MIMIC-III), and ICU mortality with heart/respiratory failure prediction (HiRID).} 
        \label{fig:MIMICX_Explain}
    \end{figure*}

\begin{figure*}
        \centering
        \begin{subfigure}[b]{0.32\textwidth}   
            \centering 
            \includegraphics[width=\textwidth]{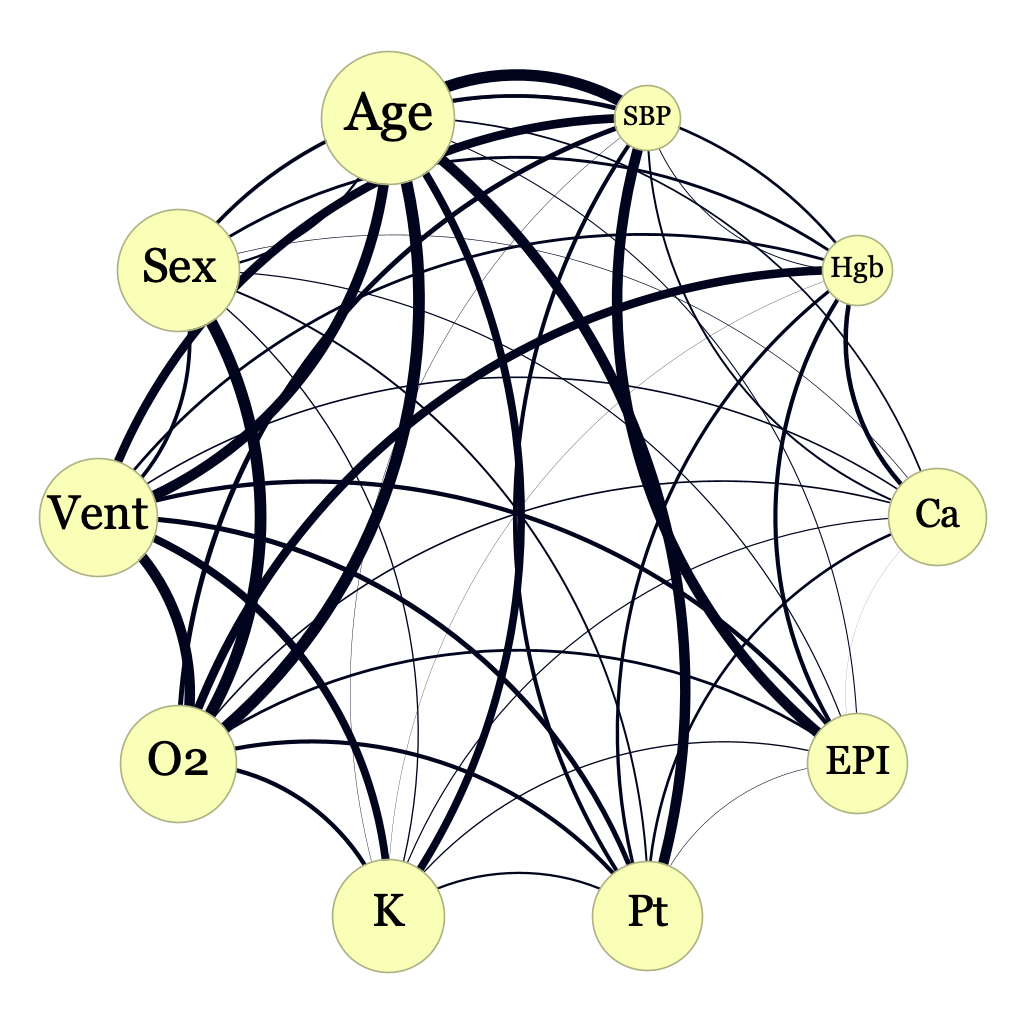}
            \caption[]%
            {{\small t=1}}    
        \end{subfigure}
        \hfill
        \begin{subfigure}[b]{0.32\textwidth}   
            \centering 
            \includegraphics[width=\textwidth]{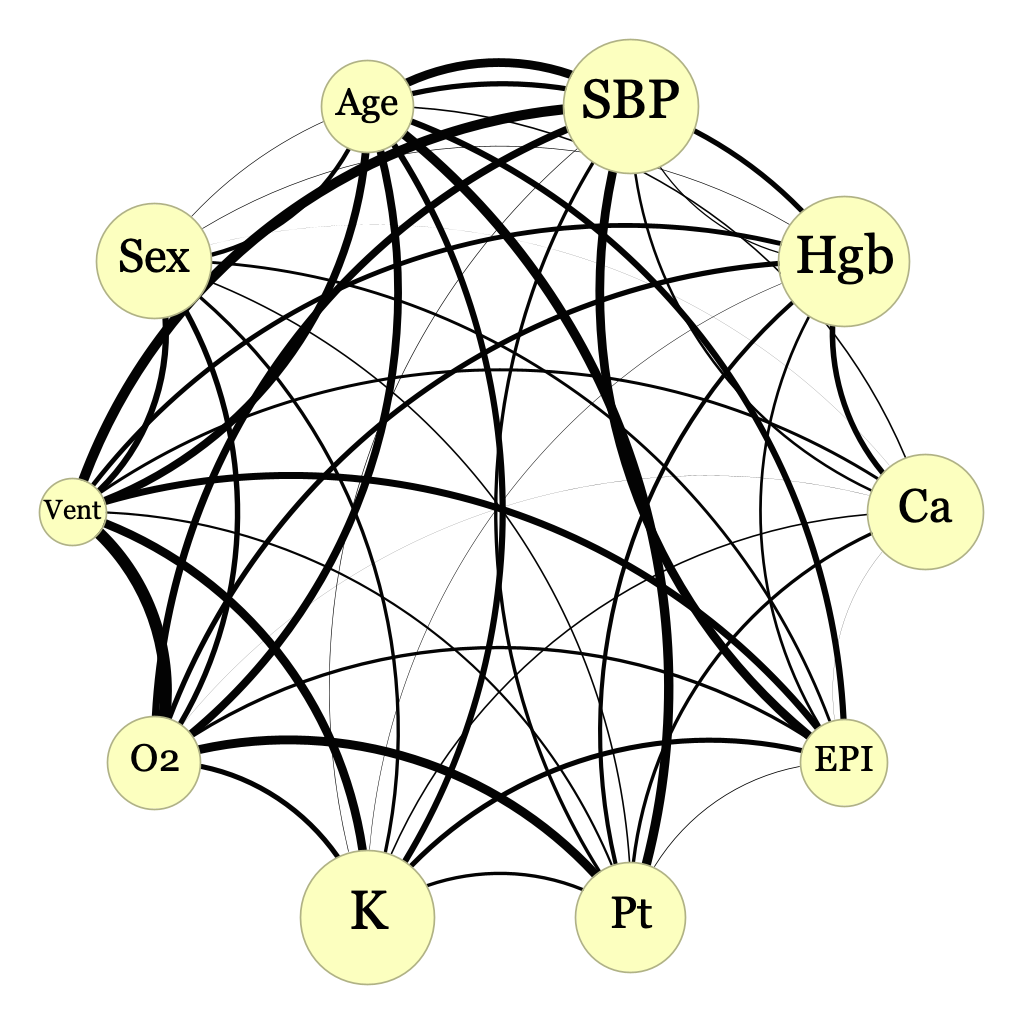}
            \caption[]%
            {{\small t=4}}
            \label{fig:mean and std of net44}
        \end{subfigure}
        \hfill
        \begin{subfigure}[b]{0.32\textwidth}   
            \centering 
            \includegraphics[width=\textwidth]{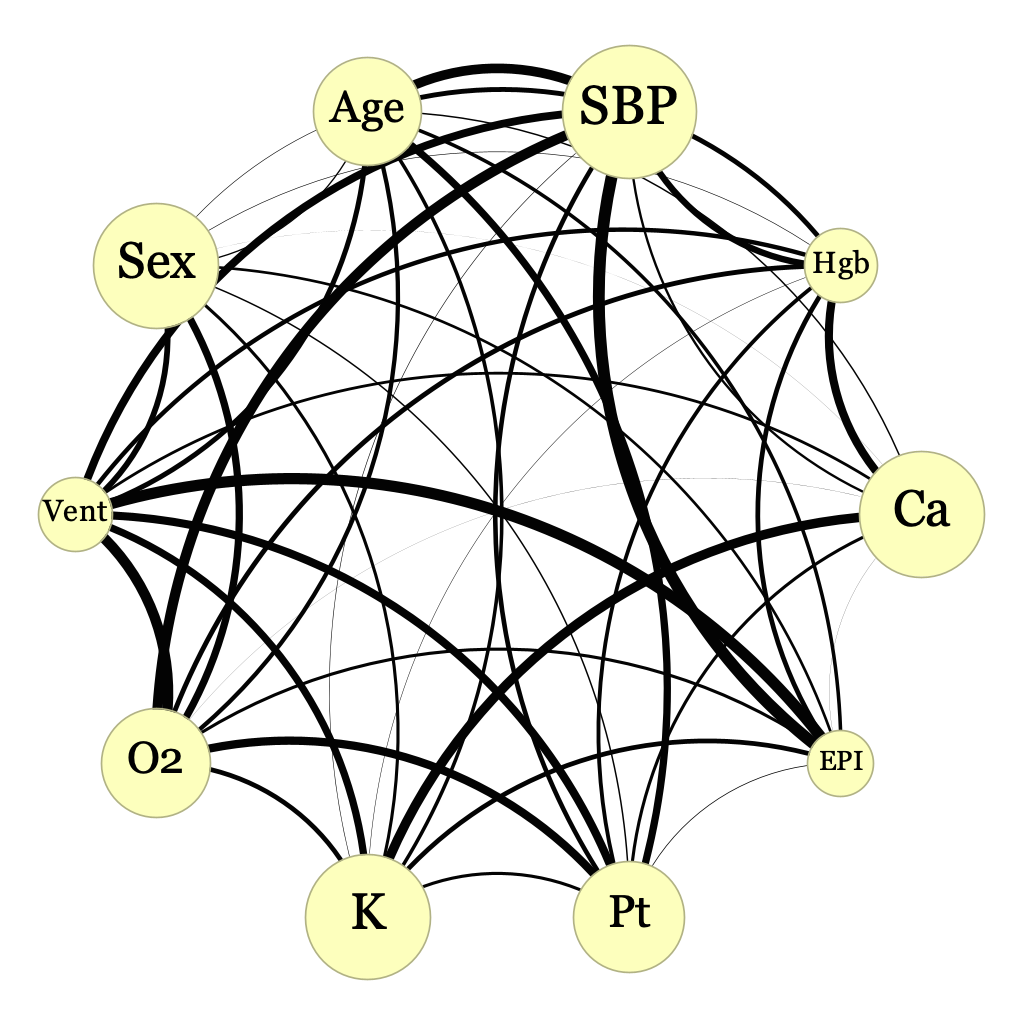}
            \caption[]%
            {{\small t=6}}
            \label{fig:mean and std of net44}
        \end{subfigure}
        \caption
        {\small Learned graph representations during DynaGraph training on the MIMIC-III dataset at two distinct timesteps: (a) $t=1$, (b) $t=4$, and (c) $t=6$, each corresponding to the first, fourth, and last 4-hour interval of a 24-hour ICU stay. The visualization highlights the evolving patterns captured by the model, including changes in individual feature importance and correlations between features. Node size corresponds to node weight magnitude, while edge darkness reflects edge weight magnitude, as derived from the interpretability matrices. These representations demonstrate the model's ability to dynamically adapt to temporal changes in the data.} 
        \label{fig:MIMICX_Explain1}
    \end{figure*}

Normalised importance scores, as described in the Methods section, are visualised in a heatmap (Figure \ref{fig:MIMICX_Explain}), providing a time-resolved representation of the significance of the features. Corresponding interpretability weights for each edge and node highlight the importance of that edge or node to the final loss computation. Larger weight magnitudes mean stronger importance. Each heatmap reveals how the importance of specific features evolves across different time steps of the prediction horizon, with each step corresponding to a segment of the time-series (e.g., 4-hour intervals for ICU datasets or 4 days in the primary care dataset). The learnt interpretability matrices, visualised at varying time points, offer granular insights into the spatio-temporal patterns captured by the model. Notably, these matrices exhibit some sparsity, reflecting the model's ability to learn meaningful, task-specific correlations rather than relying on high-level feature aggregations. The weight distributions also demonstrate variability across time steps within the same dataset, highlighting the dynamic nature of feature relationships as learned by the graph model. Figure \ref{fig:MIMICX_Explain1} illustrates the changes in important features for the patient's journey (i.e., an ICU stay). Larger nodes mean those nodes had a higher corresponding weight in the interpretability matrix (graph) and darker edges mean those edges had a higher corresponding weight in the same matrix. This behaviour shows that static risk factors, such as age and sex, tend to dominate early in the stay due to the limited temporal information available at that stage. As the stay progresses, the model increasingly incorporates dynamic vital signs, revealing correlations that align with clinical evidence on the importance of age- and sex-adjusted risk stratification in intensive care \citep{candel2022association}. This temporal adaptability underscores the model's capacity to capture evolving clinical dynamics, which is critical for tasks such as patient outcome prediction.

These visualisations improve our understanding of the model decision-making process and provide actionable insights for clinicians. The model can guide clinical decision-making and resource allocation by identifying which features become critical at different stages of patient care. A more in-depth discussion of these results and their implications for patient management is presented in the Discussion section.

\subsection{Model Evaluation }
\subsubsection{Ablation Studies}
\begin{figure*}
        \centering
        \begin{subfigure}[b]{0.47\textwidth}   
            \centering 
            \includegraphics[width=\textwidth]{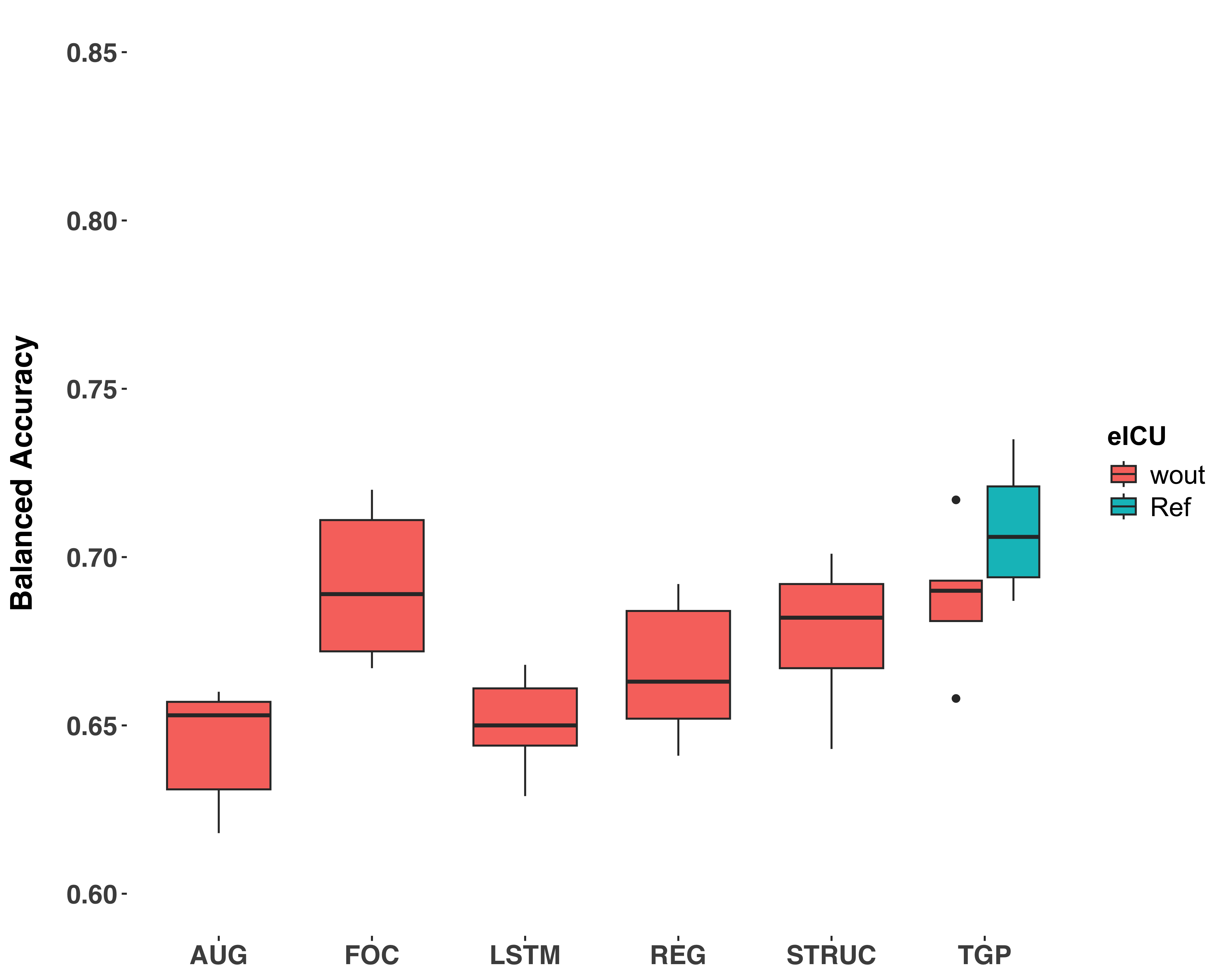}
            \caption[eICU]%
            {{\small}}    
            \label{fig:mean and std of net34}
        \end{subfigure}
        \hfill
        \begin{subfigure}[b]{0.47\textwidth}   
            \centering 
            \includegraphics[width=\textwidth]{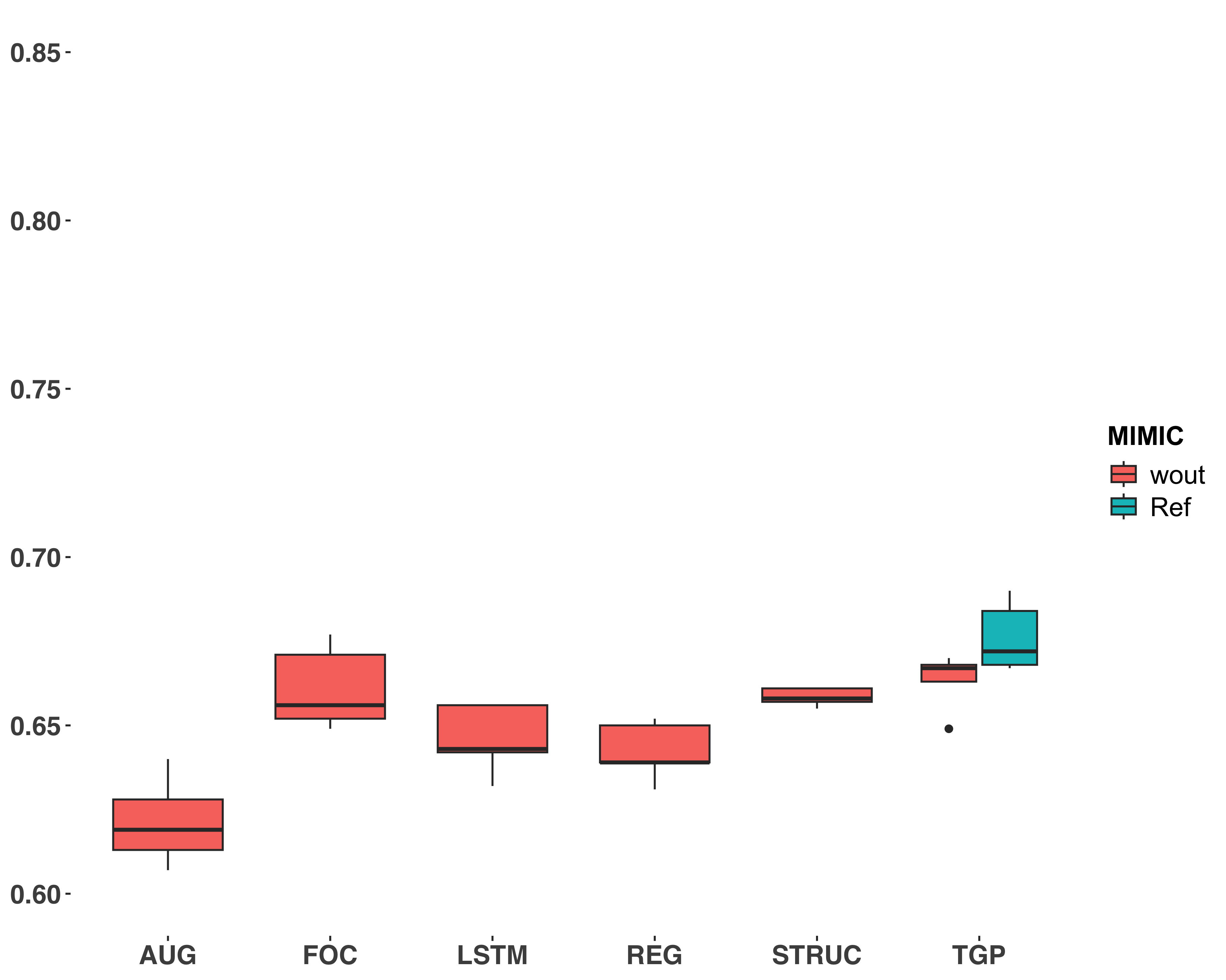}
           \caption[MIMIC-III]%
            {{\small}}
            \label{fig:mean and std of net44}
        \end{subfigure}
        \vfill
        \begin{subfigure}[b]{0.47\textwidth}   
            \centering 
            \includegraphics[width=\textwidth]{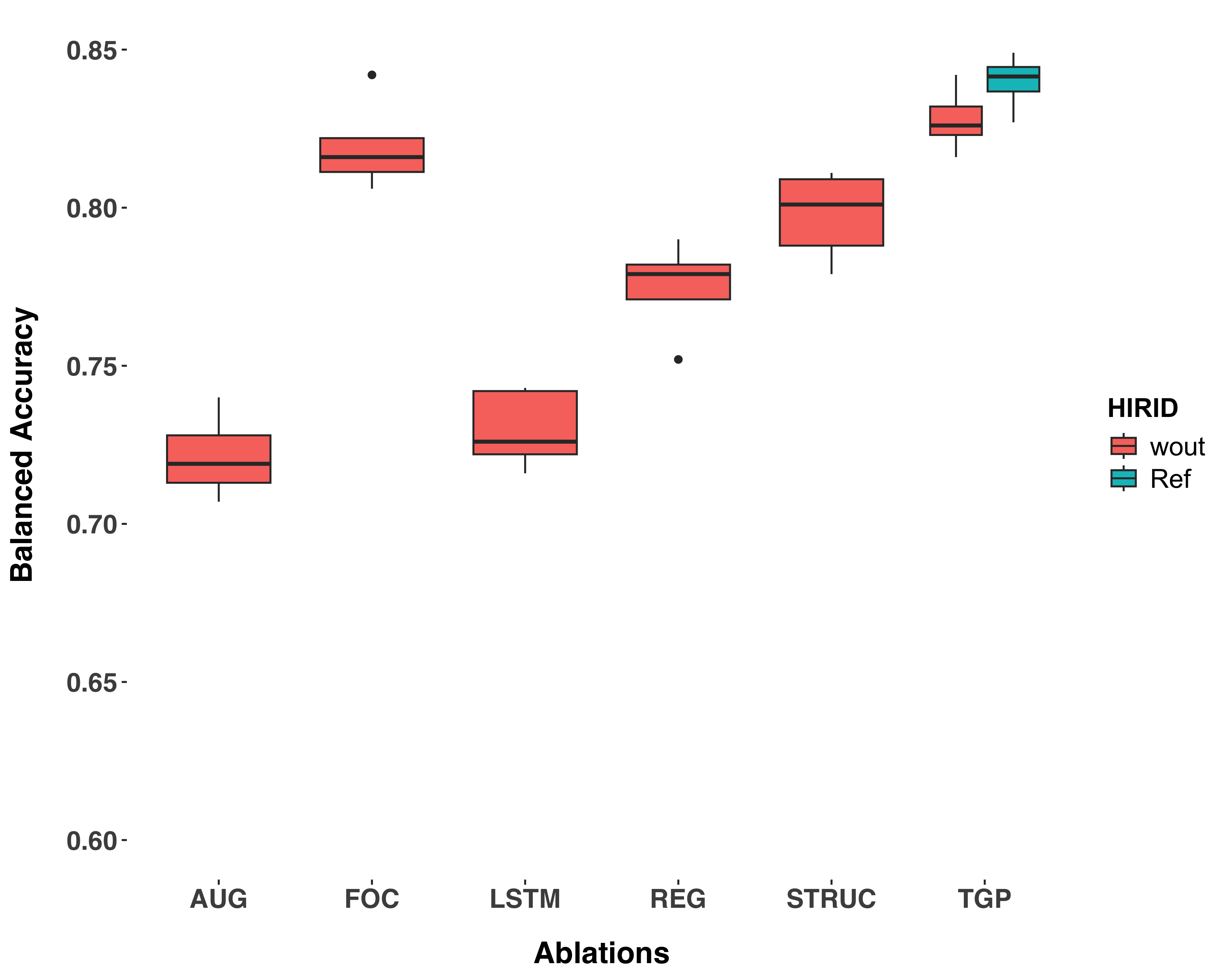}
            \caption[HiRID-ICU]%
            {{\small}}    
            \label{fig:mean and std of net34}
        \end{subfigure}
        \hfill
        \begin{subfigure}[b]{0.47\textwidth}   
            \centering 
            \includegraphics[width=\textwidth]{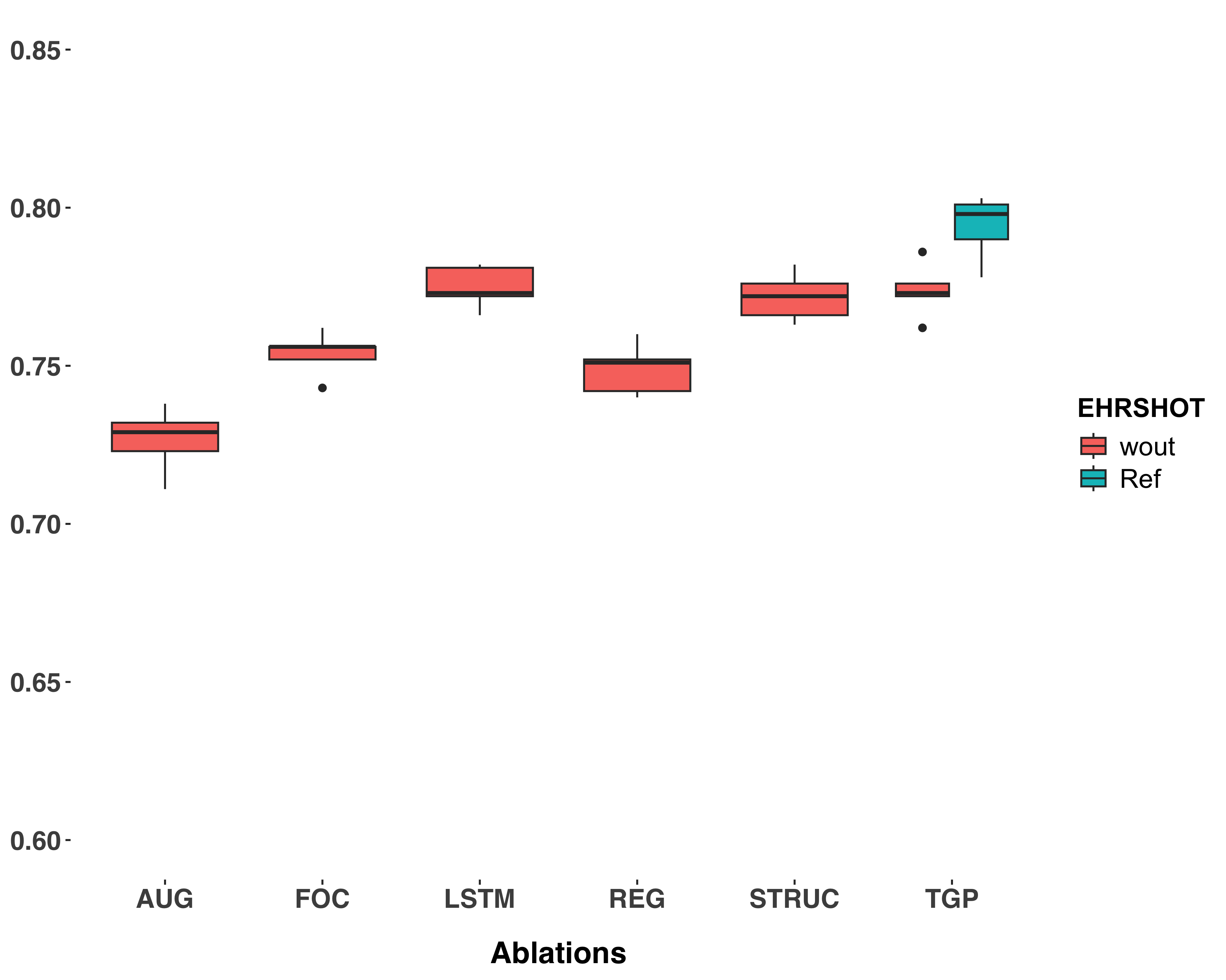}
           \caption[EHRSHOT]%
            {{\small}}
            \label{fig:mean and std of net44}
        \end{subfigure}
        \caption
        {The ablation studies for (a) eICU, (b) MIMIC-III, (c) HiRID-ICU, and (d) EHRSHOT datasets under the balanced accuracy metric. The x-axis contains the different modules excluded for the ablations, namely, AUG for graph augmentation and contrastive loss, FOC for the focal loss, LSTM for the LSTM embedding module, REG for the regularisation loss, STRUC for the structural loss, and TGP for the temporal graph pooling. The legend contains the name of the dataset used, \textit{wout} stands for the model performance without that module included, and \textit{Ref} is the best-performing model with all the modules included.} 
        \label{fig:Ablation}
    \end{figure*}

To empirically validate the contributions of individual components within the DynaGraph framework, we performed comprehensive ablation studies on the MIMIC-III, eICU, and HiRID-ICU datasets. The results presented in Figure \ref{fig:Ablation} illustrate the impact of excluding specific components, with each subfigure corresponding to a distinct ablation. Our findings demonstrate that the full DynaGraph architecture, which incorporates all components, consistently achieves superior performance across the majority of experimental settings. Notably, temporal graph pooling not only reduces computational complexity, but also enhances model performance, highlighting its dual role in efficiency and effectiveness. We specifically evaluate the impact of our proposed dynamic graph augmentation strategy, which integrates random node masking and edge perturbations with a contrastive loss objective. To assess its efficacy, we compare two variants of the model: one with the augmentation strategy and one without. The augmented variant exhibits significant improvements in balanced accuracy, demonstrating the model’s ability to learn robust representations that are invariant to irrelevant transformations while preserving sensitivity to semantically meaningful variations in the graph structure. In addition, we analyse the contributions of structural loss and focal loss through independent ablation experiments, evaluating the model performance with and without each loss term. The results indicate that both losses positively influence DynaGraph performance. The focal loss effectively addresses class imbalance, as evidenced by a notable increase in balanced accuracy. Meanwhile, structural loss ensures temporal stability in graph representations, mitigating abrupt variations between consecutive graph snapshots. To further enhance generalisation, we introduce a regularisation loss, which significantly reduces overfitting and improves performance on test sets, particularly given the large parameterisation of the model. 

\subsubsection{Out-Of-Distribution Experiments}
\begin{figure*}[t]
    \centering
    \includegraphics[width=0.8\linewidth]{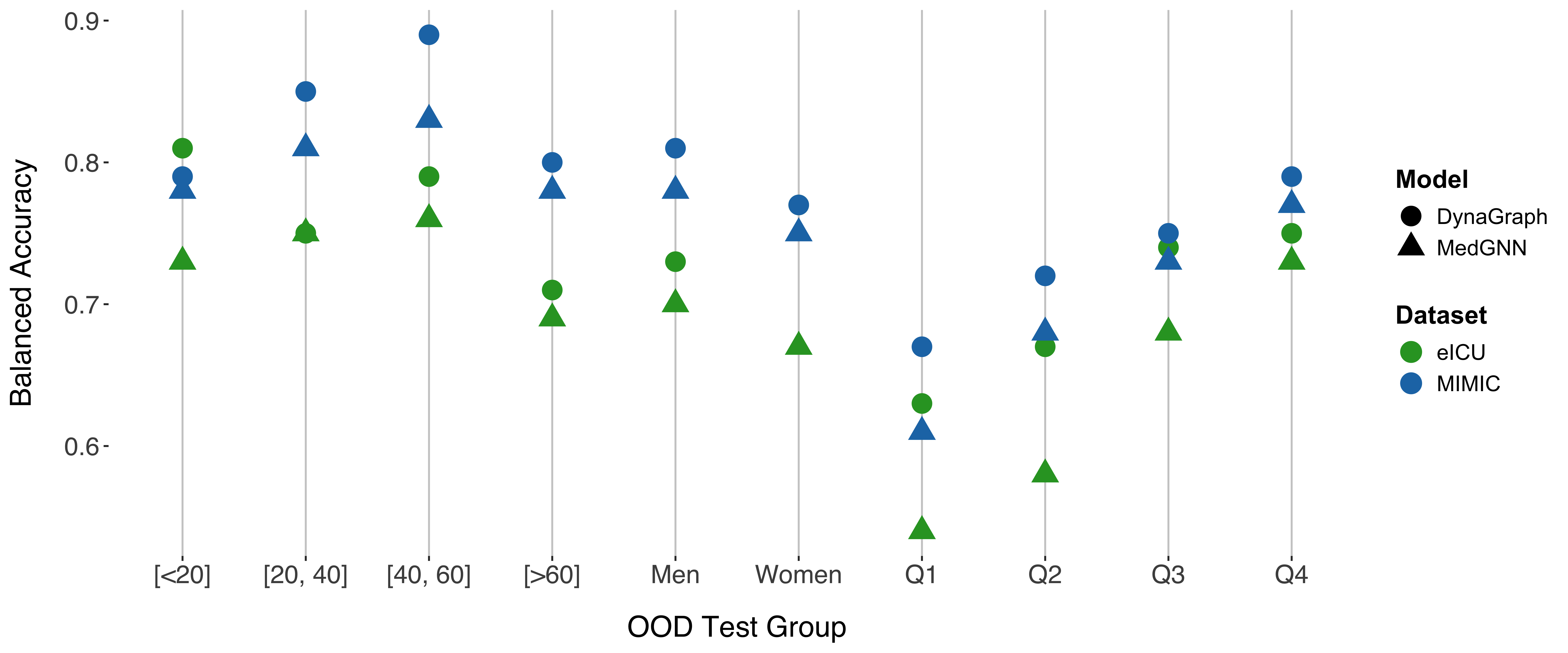}
    \caption{Generalisation performance of DynaGraph and MedGNN on out-of-distribution (OOD) settings, evaluated using balanced accuracy. OOD test sets were constructed by excluding specific subgroups like age groups: $<20$ to $>80$; gender: male and female; and time quartiles: Q1 (first 6 hours of stay) to Q4 (last 6 hours of stay) from the training data on the eICU and MIMIC-III datasets. DynaGraph demonstrates superior robustness across all OOD scenarios compared to MedGNN.}
    \label{fig:OOD}
\end{figure*}

Figure \ref{fig:OOD} compares the performance of DynaGraph with that of its closest competitor, MedGNN, in terms of balanced accuracy across various out-of-distribution (OOD) experimental settings. To evaluate robustness, we construct test sets that contain different subgroups of data (e.g., specific patient demographics or timepoints) and exclude these subgroups from the training set. Then both models are evaluated on these OOD test sets under each scenario. In all settings, DynaGraph consistently outperforms MedGNN, demonstrating superior generalisability. In particular, MedGNN exhibits a higher vulnerability to OOD samples, particularly in the early timepoint quartile and specific age groups, where its balanced accuracy degrades significantly. These results highlight DynaGraph's robustness to distribution shifts and its ability to maintain reliable performance in challenging OOD scenarios.

\section{DISCUSSION}
A common simplifying assumption in prior work on dynamic graph learning is that a static graph representation can sufficiently capture the temporal and structural information required for downstream tasks. Models such as GCNs and GATs often adopt this approach, but they fail to account for potential associations that arise when time-series data are represented by multiple graphs over time \citep{kipf2016semi, cao2020spectral}. Our ablation studies demonstrate that combining dynamic graph representations with sequential modelling techniques, such as LSTMs, significantly improves predictive performance. This improvement stems from the model's ability to jointly capture spatial and temporal patterns in multivariate time-series, as well as hidden feature codependencies over time. DynaGraph is also free from the constraining assumptions of predefined graph constructs required in other dynamic graph models as it effectively learns the graph constructions from random initialisation.

DynaGraph achieves reliable performance on the challenging task of multi-label multivariate time-series classification, outperforming both traditional time-series models and graph-based approaches. By balancing focal, contrastive, and structural losses, DynaGraph maintains robust predictive performance even in the presence of severe class imbalance across datasets. In particular, the model exhibits a substantial improvement in sensitivity, demonstrating its ability to identify positive samples in multi-label settings better. This is attributed to the class imbalance strategies embedded in the multi-loss framework.

The interpretability of DynaGraph, as illustrated in Figure \ref{fig:MIMICX_Explain}, aligns with established clinical knowledge. For example, age and sex are identified as stable predictors of cardiovascular outcomes in patients with heart attack, consistent with previous literature \citep{desai2011long}. In addition, dynamic features such as blood pressure, potassium, and calcium levels are highlighted as consistent indicators of a higher risk of complications in the ICU \citep{damluji2021mechanical, hou2024serum}. In particular, DynaGraph identifies sodium and haemoglobin levels as critical predictors closer to the time of complication events, a finding supported by medical research suggesting their relevance in at-risk patients \citep{sakr2013fluctuations, huang2016lower}. These results highlight the ability of our model to capture clinically significant risk factors and their temporal variability.

Despite these strengths, our work has several limitations. We plan to investigate interpretability weights by stratifying patient characteristics, such as age, sex, and admission criteria, to estimate potential biases and uncover group-specific associations with outcomes. There are still some 

In summary, DynaGraph outperforms state-of-the-art time-series and dynamic graph models on two real-world EHR datasets, demonstrating its ability to learn clinically meaningful patterns through its pseudo-attention interpretability framework. Future work will focus on evaluating the model in more diverse datasets and advancing the interpretability methodology further to bridge the gap between machine learning and clinical practice.

\bibliography{main}

\begin{thebibliography}{51}
\providecommand{\natexlab}[1]{#1}
\providecommand{\url}[1]{\texttt{#1}}
\expandafter\ifx\csname urlstyle\endcsname\relax
  \providecommand{\doi}[1]{doi: #1}\else
  \providecommand{\doi}{doi: \begingroup \urlstyle{rm}\Url}\fi

\bibitem[Aguiar et~al.(2022)Aguiar, Santos, Watkinson, and Zhu]{aguiar2022learning}
H.~Aguiar, M.~Santos, P.~Watkinson, and T.~Zhu.
\newblock Learning of cluster-based feature importance for electronic health record time-series.
\newblock In \emph{International conference on machine learning}, pages 161--179. PMLR, 2022.

\bibitem[Bai et~al.(2020)Bai, Yao, Li, Wang, and Wang]{bai2020adaptive}
L.~Bai, L.~Yao, C.~Li, X.~Wang, and C.~Wang.
\newblock Adaptive graph convolutional recurrent network for traffic forecasting.
\newblock \emph{Advances in neural information processing systems}, 33:\penalty0 17804--17815, 2020.

\bibitem[Baytas et~al.(2017)Baytas, Xiao, Zhang, Wang, Jain, and Zhou]{baytas2017patient}
I.~M. Baytas, C.~Xiao, X.~Zhang, F.~Wang, A.~K. Jain, and J.~Zhou.
\newblock Patient subtyping via time-aware lstm networks.
\newblock In \emph{Proceedings of the 23rd ACM SIGKDD international conference on knowledge discovery and data mining}, pages 65--74, 2017.

\bibitem[Bogaerts et~al.(2020)Bogaerts, Masegosa, Angarita-Zapata, Onieva, and Hellinckx]{bogaerts2020graph}
T.~Bogaerts, A.~D. Masegosa, J.~S. Angarita-Zapata, E.~Onieva, and P.~Hellinckx.
\newblock A graph cnn-lstm neural network for short and long-term traffic forecasting based on trajectory data.
\newblock \emph{Transportation Research Part C: Emerging Technologies}, 112:\penalty0 62--77, 2020.

\bibitem[Candel et~al.(2022)Candel, Duijzer, Gaakeer, Ter~Avest, Sir, Lameijer, Hessels, Reijnen, van Zwet, de~Jonge, et~al.]{candel2022association}
B.~G. Candel, R.~Duijzer, M.~I. Gaakeer, E.~Ter~Avest, {\"O}.~Sir, H.~Lameijer, R.~Hessels, R.~Reijnen, E.~W. van Zwet, E.~de~Jonge, et~al.
\newblock The association between vital signs and clinical outcomes in emergency department patients of different age categories.
\newblock \emph{Emergency Medicine Journal}, 39\penalty0 (12):\penalty0 903--911, 2022.

\bibitem[Cao et~al.(2020)Cao, Wang, Duan, Zhang, Zhu, Huang, Tong, Xu, Bai, Tong, et~al.]{cao2020spectral}
D.~Cao, Y.~Wang, J.~Duan, C.~Zhang, X.~Zhu, C.~Huang, Y.~Tong, B.~Xu, J.~Bai, J.~Tong, et~al.
\newblock Spectral temporal graph neural network for multivariate time-series forecasting.
\newblock \emph{Advances in neural information processing systems}, 33:\penalty0 17766--17778, 2020.

\bibitem[Choi et~al.(2016)Choi, Bahadori, Sun, Kulas, Schuetz, and Stewart]{choi2016retain}
E.~Choi, M.~T. Bahadori, J.~Sun, J.~Kulas, A.~Schuetz, and W.~Stewart.
\newblock Retain: An interpretable predictive model for healthcare using reverse time attention mechanism.
\newblock \emph{Advances in neural information processing systems}, 29, 2016.

\bibitem[Damluji et~al.(2021)Damluji, Van~Diepen, Katz, Menon, Tamis-Holland, Bakitas, Cohen, Balsam, Chikwe, American Heart Association Council on Clinical Cardiology; Council~on Arteriosclerosis, on~Cardiovascular~Surgery, Anesthesia;, on~Cardiovascular, and Nursing]{damluji2021mechanical}
A.~A. Damluji, S.~Van~Diepen, J.~N. Katz, V.~Menon, J.~E. Tamis-Holland, M.~Bakitas, M.~G. Cohen, L.~B. Balsam, J.~Chikwe, T.~American Heart Association Council on Clinical Cardiology; Council~on Arteriosclerosis, V.~B.~C. on~Cardiovascular~Surgery, Anesthesia;, C.~on~Cardiovascular, and S.~Nursing.
\newblock Mechanical complications of acute myocardial infarction: a scientific statement from the american heart association.
\newblock \emph{Circulation}, 144\penalty0 (2):\penalty0 e16--e35, 2021.

\bibitem[Desai et~al.(2011)Desai, Law, and Needham]{desai2011long}
S.~V. Desai, T.~J. Law, and D.~M. Needham.
\newblock Long-term complications of critical care.
\newblock \emph{Critical care medicine}, 39\penalty0 (2):\penalty0 371--379, 2011.

\bibitem[Duan et~al.(2022)Duan, Xu, Wang, Huang, Ren, Xu, Sun, and Wang]{duan2022multivariate}
Z.~Duan, H.~Xu, Y.~Wang, Y.~Huang, A.~Ren, Z.~Xu, Y.~Sun, and W.~Wang.
\newblock Multivariate time-series classification with hierarchical variational graph pooling.
\newblock \emph{Neural Networks}, 154:\penalty0 481--490, 2022.

\bibitem[Duffy et~al.(2022)Duffy, Cheng, Yuan, He, Kwan, Shun-Shin, Alexander, Ebinger, Lungren, Rader, et~al.]{duffy2022high}
G.~Duffy, P.~P. Cheng, N.~Yuan, B.~He, A.~C. Kwan, M.~J. Shun-Shin, K.~M. Alexander, J.~Ebinger, M.~P. Lungren, F.~Rader, et~al.
\newblock High-throughput precision phenotyping of left ventricular hypertrophy with cardiovascular deep learning.
\newblock \emph{JAMA cardiology}, 7\penalty0 (4):\penalty0 386--395, 2022.

\bibitem[Elbadawi et~al.(2019)Elbadawi, Elgendy, Mahmoud, Barakat, Mentias, Mohamed, Ogunbayo, Megaly, Saad, Omer, et~al.]{elbadawi2019temporal}
A.~Elbadawi, I.~Y. Elgendy, K.~Mahmoud, A.~F. Barakat, A.~Mentias, A.~H. Mohamed, G.~O. Ogunbayo, M.~Megaly, M.~Saad, M.~A. Omer, et~al.
\newblock Temporal trends and outcomes of mechanical complications in patients with acute myocardial infarction.
\newblock \emph{JACC: Cardiovascular Interventions}, 12\penalty0 (18):\penalty0 1825--1836, 2019.

\bibitem[Fan et~al.(2025)Fan, Fei, Guo, Yi, Song, Xiang, Ye, and Li]{fan2025medgnn}
W.~Fan, J.~Fei, D.~Guo, K.~Yi, X.~Song, H.~Xiang, H.~Ye, and M.~Li.
\newblock Medgnn: Towards multi-resolution spatiotemporal graph learning for medical time series classification.
\newblock \emph{arXiv preprint arXiv:2502.04515}, 2025.

\bibitem[Hamilton et~al.(2017)Hamilton, Ying, and Leskovec]{hamilton2017inductive}
W.~Hamilton, Z.~Ying, and J.~Leskovec.
\newblock Inductive representation learning on large graphs.
\newblock \emph{Advances in neural information processing systems}, 30, 2017.

\bibitem[Han et~al.(2021)Han, Du, Sun, Fu, Lv, and Xiong]{han2021dynamic}
L.~Han, B.~Du, L.~Sun, Y.~Fu, Y.~Lv, and H.~Xiong.
\newblock Dynamic and multi-faceted spatio-temporal deep learning for traffic speed forecasting.
\newblock In \emph{Proceedings of the 27th ACM SIGKDD conference on knowledge discovery \& data mining}, pages 547--555, 2021.

\bibitem[Harutyunyan et~al.(2019)Harutyunyan, Khachatrian, Kale, Ver~Steeg, and Galstyan]{harutyunyan2019multitask}
H.~Harutyunyan, H.~Khachatrian, D.~C. Kale, G.~Ver~Steeg, and A.~Galstyan.
\newblock Multitask learning and benchmarking with clinical time series data.
\newblock \emph{Scientific data}, 6\penalty0 (1):\penalty0 96, 2019.

\bibitem[Hou et~al.(2024)Hou, Huang, Zeng, Wu, and Zhang]{hou2024serum}
J.~Hou, Z.~Huang, W.~Zeng, Z.~Wu, and L.~Zhang.
\newblock Serum calcium is associated with sudden cardiac arrest in stroke patients from icu: a multicenter retrospective study based on the eicu collaborative research database.
\newblock \emph{Scientific Reports}, 14\penalty0 (1):\penalty0 1700, 2024.

\bibitem[Huang et~al.(2023)Huang, Jiang, Rao, Zhang, Han, Zhang, Wang, He, Xu, Zhao, et~al.]{huang2023benchtemp}
Q.~Huang, J.~Jiang, X.~S. Rao, C.~Zhang, Z.~Han, Z.~Zhang, X.~Wang, Y.~He, Q.~Xu, Y.~Zhao, et~al.
\newblock Benchtemp: A general benchmark for evaluating temporal graph neural networks.
\newblock \emph{arXiv preprint arXiv:2308.16385}, 2023.

\bibitem[Huang and Hu(2016)]{huang2016lower}
Y.-L. Huang and Z.-D. Hu.
\newblock Lower mean corpuscular hemoglobin concentration is associated with poorer outcomes in intensive care unit admitted patients with acute myocardial infarction.
\newblock \emph{Annals of translational medicine}, 4\penalty0 (10), 2016.

\bibitem[Hyland et~al.(2020)Hyland, Faltys, H{\"u}ser, Lyu, Gumbsch, Esteban, Bock, Horn, Moor, Rieck, et~al.]{hyland2020early}
S.~L. Hyland, M.~Faltys, M.~H{\"u}ser, X.~Lyu, T.~Gumbsch, C.~Esteban, C.~Bock, M.~Horn, M.~Moor, B.~Rieck, et~al.
\newblock Early prediction of circulatory failure in the intensive care unit using machine learning.
\newblock \emph{Nature medicine}, 26\penalty0 (3):\penalty0 364--373, 2020.

\bibitem[Johnson et~al.(2016)Johnson, Pollard, Shen, Lehman, Feng, Ghassemi, Moody, Szolovits, Anthony~Celi, and Mark]{johnson2016mimic}
A.~E. Johnson, T.~J. Pollard, L.~Shen, L.-w.~H. Lehman, M.~Feng, M.~Ghassemi, B.~Moody, P.~Szolovits, L.~Anthony~Celi, and R.~G. Mark.
\newblock Mimic-iii, a freely accessible critical care database.
\newblock \emph{Scientific data}, 3\penalty0 (1):\penalty0 1--9, 2016.

\bibitem[Kipf and Welling(2016)]{kipf2016semi}
T.~N. Kipf and M.~Welling.
\newblock Semi-supervised classification with graph convolutional networks.
\newblock \emph{arXiv preprint arXiv:1609.02907}, 2016.

\bibitem[Kok et~al.(2020)Kok, Jahmunah, Oh, Zhou, Gururajan, Tao, Cheong, Gururajan, Molinari, and Acharya]{kok2020automated}
C.~Kok, V.~Jahmunah, S.~L. Oh, X.~Zhou, R.~Gururajan, X.~Tao, K.~H. Cheong, R.~Gururajan, F.~Molinari, and U.~R. Acharya.
\newblock Automated prediction of sepsis using temporal convolutional network.
\newblock \emph{Computers in Biology and Medicine}, 127:\penalty0 103957, 2020.

\bibitem[Lauritsen et~al.(2020)Lauritsen, Kristensen, Olsen, Larsen, Lauritsen, J{\o}rgensen, Lange, and Thiesson]{lauritsen2020explainable}
S.~M. Lauritsen, M.~Kristensen, M.~V. Olsen, M.~S. Larsen, K.~M. Lauritsen, M.~J. J{\o}rgensen, J.~Lange, and B.~Thiesson.
\newblock Explainable artificial intelligence model to predict acute critical illness from electronic health records.
\newblock \emph{Nature communications}, 11\penalty0 (1):\penalty0 3852, 2020.

\bibitem[Lee and Van Der~Schaar(2020)]{lee2020temporal}
C.~Lee and M.~Van Der~Schaar.
\newblock Temporal phenotyping using deep predictive clustering of disease progression.
\newblock In \emph{International Conference on Machine Learning}, pages 5767--5777. PMLR, 2020.

\bibitem[Li and Zhu(2021)]{li2021spatial}
M.~Li and Z.~Zhu.
\newblock Spatial-temporal fusion graph neural networks for traffic flow forecasting.
\newblock In \emph{Proceedings of the AAAI conference on artificial intelligence}, volume~35, pages 4189--4196, 2021.

\bibitem[Li et~al.(2020)Li, Rao, Solares, Hassaine, Ramakrishnan, Canoy, Zhu, Rahimi, and Salimi-Khorshidi]{li2020behrt}
Y.~Li, S.~Rao, J.~R.~A. Solares, A.~Hassaine, R.~Ramakrishnan, D.~Canoy, Y.~Zhu, K.~Rahimi, and G.~Salimi-Khorshidi.
\newblock Behrt: transformer for electronic health records.
\newblock \emph{Scientific reports}, 10\penalty0 (1):\penalty0 7155, 2020.

\bibitem[Liu et~al.(2023)Liu, Liu, Yang, Liang, Wang, Cui, and Gu]{liu2023todynet}
H.~Liu, X.~Liu, D.~Yang, Z.~Liang, H.~Wang, Y.~Cui, and J.~Gu.
\newblock Todynet: Temporal dynamic graph neural network for multivariate time series classification.
\newblock \emph{arXiv preprint arXiv:2304.05078}, 2023.

\bibitem[Luo et~al.(2023)Luo, Gong, and Li]{luo2023pt3}
R.~Luo, M.~Gong, and C.~Li.
\newblock Pt3: A transformer-based model for sepsis death risk prediction via vital signs time series.
\newblock In \emph{2023 International Joint Conference on Neural Networks (IJCNN)}, pages 1--9. IEEE, 2023.

\bibitem[Pollard et~al.(2018)Pollard, Johnson, Raffa, Celi, Mark, and Badawi]{pollard2018eicu}
T.~J. Pollard, A.~E. Johnson, J.~D. Raffa, L.~A. Celi, R.~G. Mark, and O.~Badawi.
\newblock The eicu collaborative research database, a freely available multi-center database for critical care research.
\newblock \emph{Scientific data}, 5\penalty0 (1):\penalty0 1--13, 2018.

\bibitem[Rafiei et~al.(2021)Rafiei, Rezaee, Hajati, Gheisari, and Golzan]{rafiei2021ssp}
A.~Rafiei, A.~Rezaee, F.~Hajati, S.~Gheisari, and M.~Golzan.
\newblock Ssp: Early prediction of sepsis using fully connected lstm-cnn model.
\newblock \emph{Computers in biology and medicine}, 128:\penalty0 104110, 2021.

\bibitem[Rocheteau et~al.(2021)Rocheteau, Tong, Veli{\v{c}}kovi{\'c}, Lane, and Li{\`o}]{rocheteau2021predicting}
E.~Rocheteau, C.~Tong, P.~Veli{\v{c}}kovi{\'c}, N.~Lane, and P.~Li{\`o}.
\newblock Predicting patient outcomes with graph representation learning.
\newblock \emph{arXiv preprint arXiv:2101.03940}, 2021.

\bibitem[Sakr et~al.(2013)Sakr, Rother, Ferreira, Ewald, D{\"u}nisch, Riedemmann, and Reinhart]{sakr2013fluctuations}
Y.~Sakr, S.~Rother, A.~M.~P. Ferreira, C.~Ewald, P.~D{\"u}nisch, N.~Riedemmann, and K.~Reinhart.
\newblock Fluctuations in serum sodium level are associated with an increased risk of death in surgical icu patients.
\newblock \emph{Critical care medicine}, 41\penalty0 (1):\penalty0 133--142, 2013.

\bibitem[Shamout et~al.(2020)Shamout, Zhu, and Clifton]{shamout2020machine}
F.~Shamout, T.~Zhu, and D.~A. Clifton.
\newblock Machine learning for clinical outcome prediction.
\newblock \emph{IEEE reviews in Biomedical Engineering}, 14:\penalty0 116--126, 2020.

\bibitem[Sheikhalishahi et~al.(2019)Sheikhalishahi, Balaraman, and Osmani]{sheikhalishahi2019benchmarking}
S.~Sheikhalishahi, V.~Balaraman, and V.~Osmani.
\newblock Benchmarking machine learning models on eicu critical care dataset.
\newblock \emph{arXiv preprint arXiv:1910.00964}, 2019.

\bibitem[Siontis et~al.(2021)Siontis, Noseworthy, Attia, and Friedman]{siontis2021artificial}
K.~C. Siontis, P.~A. Noseworthy, Z.~I. Attia, and P.~A. Friedman.
\newblock Artificial intelligence-enhanced electrocardiography in cardiovascular disease management.
\newblock \emph{Nature Reviews Cardiology}, 18\penalty0 (7):\penalty0 465--478, 2021.

\bibitem[Smith et~al.(2014)Smith, Chiovaro, O’Neil, Kansagara, Qui{\~n}ones, Freeman, Motu’apuaka, and Slatore]{smith2014early}
M.~B. Smith, J.~C. Chiovaro, M.~O’Neil, D.~Kansagara, A.~R. Qui{\~n}ones, M.~Freeman, M.~L. Motu’apuaka, and C.~G. Slatore.
\newblock Early warning system scores for clinical deterioration in hospitalized patients: a systematic review.
\newblock \emph{Annals of the American Thoracic Society}, 11\penalty0 (9):\penalty0 1454--1465, 2014.

\bibitem[Tan et~al.(2020)Tan, Ye, Yang, Liu, Ma, Yip, Wong, and Yuen]{tan2020data}
Q.~Tan, M.~Ye, B.~Yang, S.~Liu, A.~J. Ma, T.~C.-F. Yip, G.~L.-H. Wong, and P.~Yuen.
\newblock Data-gru: Dual-attention time-aware gated recurrent unit for irregular multivariate time series.
\newblock In \emph{Proceedings of the AAAI Conference on Artificial Intelligence}, volume~34, pages 930--937, 2020.

\bibitem[Veli{\v{c}}kovi{\'c} et~al.(2017)Veli{\v{c}}kovi{\'c}, Cucurull, Casanova, Romero, Lio, and Bengio]{velivckovic2017graph}
P.~Veli{\v{c}}kovi{\'c}, G.~Cucurull, A.~Casanova, A.~Romero, P.~Lio, and Y.~Bengio.
\newblock Graph attention networks.
\newblock \emph{arXiv preprint arXiv:1710.10903}, 2017.

\bibitem[Wang et~al.(2023)Wang, Chen, Jin, Ren, Wang, Cao, and Xia]{wang2023bit}
Q.~Wang, G.~Chen, X.~Jin, S.~Ren, G.~Wang, L.~Cao, and Y.~Xia.
\newblock Bit-mac: Mortality prediction by bidirectional time and multi-feature attention coupled network on multivariate irregular time series.
\newblock \emph{Computers in Biology and Medicine}, 155:\penalty0 106586, 2023.

\bibitem[Wang and Liu(2020)]{wang2020tagnet}
S.~Wang and J.~Liu.
\newblock Tagnet: Temporal aware graph convolution network for clinical information extraction.
\newblock In \emph{2020 IEEE International Conference on Bioinformatics and Biomedicine (BIBM)}, pages 2105--2108. IEEE, 2020.

\bibitem[Wang et~al.(2020)Wang, McDermott, Chauhan, Ghassemi, Hughes, and Naumann]{wang2020mimic}
S.~Wang, M.~B. McDermott, G.~Chauhan, M.~Ghassemi, M.~C. Hughes, and T.~Naumann.
\newblock Mimic-extract: A data extraction, preprocessing, and representation pipeline for mimic-iii.
\newblock In \emph{Proceedings of the ACM conference on health, inference, and learning}, pages 222--235, 2020.

\bibitem[Wang and Aste(2022)]{wang2022sparsification}
Y.~Wang and T.~Aste.
\newblock Sparsification and filtering for spatial-temporal gnn in multivariate time-series.
\newblock \emph{arXiv preprint arXiv:2203.03991}, 2022.

\bibitem[Wang et~al.(2022)Wang, Zhao, Callcut, and Petzold]{wang2022integrating}
Y.~Wang, Y.~Zhao, R.~Callcut, and L.~Petzold.
\newblock Integrating physiological time series and clinical notes with transformer for early prediction of sepsis.
\newblock \emph{arXiv preprint arXiv:2203.14469}, 2022.

\bibitem[Wang et~al.(2024)Wang, Huang, Li, Yan, and Zhang]{wang2024medformer}
Y.~Wang, N.~Huang, T.~Li, Y.~Yan, and X.~Zhang.
\newblock Medformer: A multi-granularity patching transformer for medical time-series classification.
\newblock \emph{arXiv preprint arXiv:2405.19363}, 2024.

\bibitem[Wornow et~al.(2024)Wornow, Thapa, Steinberg, Fries, and Shah]{wornow2024ehrshot}
M.~Wornow, R.~Thapa, E.~Steinberg, J.~Fries, and N.~Shah.
\newblock Ehrshot: An ehr benchmark for few-shot evaluation of foundation models.
\newblock \emph{Advances in Neural Information Processing Systems}, 36, 2024.

\bibitem[Wu et~al.(2020)Wu, Pan, Long, Jiang, Chang, and Zhang]{wu2020connecting}
Z.~Wu, S.~Pan, G.~Long, J.~Jiang, X.~Chang, and C.~Zhang.
\newblock Connecting the dots: Multivariate time series forecasting with graph neural networks.
\newblock In \emph{Proceedings of the 26th ACM SIGKDD international conference on knowledge discovery \& data mining}, pages 753--763, 2020.

\bibitem[Xu et~al.(2021)Xu, Duan, Wang, Feng, Chen, Zhang, and Xu]{xu2021graph}
H.~Xu, Z.~Duan, Y.~Wang, J.~Feng, R.~Chen, Q.~Zhang, and Z.~Xu.
\newblock Graph partitioning and graph neural network based hierarchical graph matching for graph similarity computation.
\newblock \emph{Neurocomputing}, 439:\penalty0 348--362, 2021.

\bibitem[Y{\`e}che et~al.(2021)Y{\`e}che, Kuznetsova, Zimmermann, H{\"u}ser, Lyu, Faltys, and R{\"a}tsch]{yeche2021hirid}
H.~Y{\`e}che, R.~Kuznetsova, M.~Zimmermann, M.~H{\"u}ser, X.~Lyu, M.~Faltys, and G.~R{\"a}tsch.
\newblock Hirid-icu-benchmark--a comprehensive machine learning benchmark on high-resolution icu data.
\newblock \emph{arXiv preprint arXiv:2111.08536}, 2021.

\bibitem[Zha et~al.(2022)Zha, Lai, Zhou, and Hu]{zha2022towards}
D.~Zha, K.-H. Lai, K.~Zhou, and X.~Hu.
\newblock Towards similarity-aware time-series classification.
\newblock In \emph{Proceedings of the 2022 SIAM International Conference on Data Mining (SDM)}, pages 199--207. SIAM, 2022.

\bibitem[Zhou et~al.(2020)Zhou, Cui, Hu, Zhang, Yang, Liu, Wang, Li, and Sun]{zhou2020graph}
J.~Zhou, G.~Cui, S.~Hu, Z.~Zhang, C.~Yang, Z.~Liu, L.~Wang, C.~Li, and M.~Sun.
\newblock Graph neural networks: A review of methods and applications.
\newblock \emph{AI open}, 1:\penalty0 57--81, 2020.

\end{thebibliography}

\end{document}


%

%

\onecolumn
\aistatstitle{Supplementary Materials}

\section{Code and Data}
\subsection{Data Description}

The pre-processing pipeline for MIMIC-III was based on MIMIC-Extract and the eICU was based in part on work done by Rocheteau et al. \citep{wang2020mimic, rocheteau2021temporal}. Since we are working with custom labels for complications and outcomes, some parts of the label extraction component were changed. We used the imputation as suggested by the pipeline. Data extract samples will be made available in the GitHub repository, but if readers wish to build on these results, we recommend following the above sources for data extraction.

For the time-series variables, we use forward filling as clinicians in practice would only consider the last recorded measurement. If the first set of measurements is missing for some time-varying features, instead of dropping those features or patients, we backward-fill from the closest measurement in the future. The time-series features were resampled to 1-hour intervals except in the case of HiRID which was 5 minutes. We considered only observations collected up to 24 hours before the registered outcome. Patient admissions were randomly split into train, validation and test sets (8:1:1). Details of the features included can be found in Supplementary Tables 1, 2, and 3. 

Since the data contains de-identified patient electronic health records data, it can only be obtained given the ethical review of the proposed analysis on the PhysioNet page. Some certification of training modules is also required for access. We have cited the sources for the datasets in the text accordingly under Data. Consent for data use has been obtained by the providers, de-identification and licencing are in line with HIPAA requirements and compatible with the research conducted which has passed ethical review and certification for data access. Due to potential re-identification risks, the data was not shared outside of the organisation's systems, and the sample data provided in the repository is a feature subset without IDs.

The most relevant feature distributions for eICU and MIMIC-III can be found in Table \ref{tab:features}.

\begin{table}[htp]
    \small
    \centering
    \caption{Summary of demographics and variables used for external validation across training and testing datasets. MIMIC-IV has been used separately as an external validation source with the entire dataset used as a test set.}
    \begin{tabularx}{0.67\textwidth}{l
                                    r
                                    r
                                    r
                                    c
                                 }
         \toprule
        \bf{Attributes} & \multicolumn{2}{c}{eICU (N = 1,433)} & \multicolumn{2}{c}{MIMIC-III (N = 17,279)}  \\
            \midrule
          Age (mean $\pm$ SD)  & 67.2 ($\pm$ 12.42) && 74.1 ($\pm$ 13.41)\\
          \addlinespace[0.05cm]
          Sex (male) & 924 (64.51$\%$) && 9,866 (57.13$\%$)\\
          \addlinespace[0.05cm]
          Ethnicity (Caucasian) & 1,108 (77.31$\%$) && 12,287 (71.11$\%$)\\
          \addlinespace[0.05cm]
          Ethnicity (African American) & 152 (10.61$\%$) && 1,382 (8.02$\%$)\\
          \addlinespace[0.05cm]
          Ethnicity (Hispanic) & 53 (3.72$\%$) && 518 (3.08$\%$)\\
          \addlinespace[0.05cm]
          Ethnicity (Asian) & 23 (1.63$\%$) && 346 (2.04$\%$)\\
          \addlinespace[0.05cm]
          Lactate  & 2.5 ($\pm$ 2.30) && 2.01 ($\pm$ 1.52)\\
          \addlinespace[0.05cm]
          SBP  & 120.0 ($\pm$ 16.33) && 126.32 ($\pm$ 18.81)\\
          \addlinespace[0.05cm]
          Glucose  & 147.3 ($\pm$ 56.72) && 136.50 ($\pm$ 49.30)\\
          \addlinespace[0.05cm]
          WBC  & 15.1 ($\pm$ 9.31) && 10.61 ($\pm$ 7.42)\\
          \addlinespace[0.05cm]
          RDW & 15.0 ($\pm$ 2.04) && 14.43 ($\pm$ 2.11)\\
          \addlinespace[0.05cm]
          Urea Nitrogen  & 22.8 ($\pm$ 13.41) && 22.8 ($\pm$ 17.04)\\
          \addlinespace[0.05cm]
          Bicarbonate & 24.8 ($\pm$ 4.40) && 23.3 ($\pm$ 3.10)\\
          \addlinespace[0.05cm]
          Mortality & 172 (12.00$\%$) && 1,667 (9.65$\%$)\\
            \bottomrule
    \end{tabularx}
    \label{tab:features}
\end{table}

\begin{table*} [th]
\caption{Features extracted from the MIMIC-III database. The features include demographic data collected for all patients, ICU unit-specific information like the type of unit, hospital information, vital signs, and biochemical measurements.}
\small
\centering
\begin{tabular}{@{}lccc@{}}
\toprule
\textbf{Static Variables} &  & &\\
\addlinespace[0.05cm]
\midrule
    \textit{Feature} & \textit{Type} & \textit{Feature} & \textit{Type}\\
    \midrule

Sex & binary & Admission Type & categorical\\
\addlinespace[0.05cm]
Age & integer & Insurance & categorical \\
\addlinespace[0.05cm]
ICU Type & categorical & Ethnicity & categorical \\
\midrule
\textbf{Time-series Variables} &  & &\\
\addlinespace[0.05cm]
\midrule
    \textit{Feature} & \textit{Type} & \textit{Feature} & \textit{Type}\\
    \midrule 
Anion Gap & continuous & WBC & continuous\\
\addlinespace[0.05cm]
Weight & continuous  & Temperature & continuous\\
\addlinespace[0.05cm]
SBP & continuous & DBP & continuous\\
\addlinespace[0.05cm]
Sodium & continuous & Respiratory Rate & continuous\\
\addlinespace[0.05cm]
RBC & continuous & Prothrombin Time PT & continuous\\
\addlinespace[0.05cm]
Prothrombin Time INR & continuous & Potassium & continuous\\
\addlinespace[0.05cm]
Platelets & categorical & Phosphorous & continuous\\
\addlinespace[0.05cm]
Phosphate & continuous & Partial Thromboplastin Time & continuous\\
\addlinespace[0.05cm]
Oxygen Saturation & continuous & MCGC & continuous\\
\addlinespace[0.05cm]
Magnesium & continuous & Hemoglobin & continuous\\
\addlinespace[0.05cm]
Hematocrit & continuous & Heart Rate & continuous\\
\addlinespace[0.05cm]
Glucose & continuous & Chloride & continuous\\
\addlinespace[0.05cm]
Creatinine & continuous & Calcium & continuous\\
\addlinespace[0.05cm]
BUN & continuous & Bicarbonate & continuous\\
\addlinespace[0.05cm]
Vent & binary & Vaso & binary\\
\addlinespace[0.05cm]
Adenosine & binary & Dobutamine & binary\\
\addlinespace[0.05cm]
Dopamine & binary & Epinephrine & binary\\
\addlinespace[0.05cm]
Isuprel & binary & Milrinone & binary\\
\addlinespace[0.05cm]
Norepinephrine & binary & Phenylepinephrine & binary\\
\addlinespace[0.05cm]
Vasopressin & binary & Colloid & binary\\
\addlinespace[0.05cm]
Crystalloid & binary & Intervention Duration & binary\\
  \bottomrule                          
\end{tabular}
\label{tab:features2}
\end{table*}

\begin{table*}[t]
\caption{Features extracted from the eICU database. The features include demographic data collected for all patients, ICU unit-specific information like type and number of beds, hospital information like regional location and teaching status, vital signs including respiratory rate and blood pressure, and biochemical measurements including troponin and levels of potassium and protein in the blood.}
\small
\centering
\begin{tabular}{@{}lccc@{}}
\toprule
    \textbf{Feature} & \textbf{Type} & \textbf{Feature} & \textbf{Type}\\
    \midrule
Sex & binary & Unit Stay Type & categorical\\
\addlinespace[0.05cm]
Age & integer & Num Beds Category & categorical \\
\addlinespace[0.05cm]
Height & continuous & Region & categorical\\
\addlinespace[0.05cm]
Weight & continuous &Teaching Status & binary\\
\addlinespace[0.05cm]
Ethnicity & categorical & Physician Speciality & categorical\\
\addlinespace[0.05cm]
Unit Type & categorical & Unit Type & categorical\\
\addlinespace[0.05cm]
Unit Admit Source & categorical &  Mechanical Ventilation & binary\\
\addlinespace[0.05cm]
Unit Visit Number & categorical & &\\
\addlinespace[0.05cm]
\midrule
\textbf{Time-series Variables} &  & &\\
\addlinespace[0.05cm]
\midrule
    \textit{Feature} & \textit{Type} & \textit{Feature} & \textit{Type}\\
    \midrule 
&& Base Excess & continuous \\
\addlinespace[0.05cm]
-basos & continuous & FiO2 & continuous\\
\addlinespace[0.05cm]
-eos & continuous & HCO3 & continuous\\
\addlinespace[0.05cm]
-monos & continuous & Hct & continuous\\
\addlinespace[0.05cm]
 -polys & continuous & Hgb & continuous\\
\addlinespace[0.05cm]
ALT & continuous & MCH & continuous\\
\addlinespace[0.05cm]
AST & continuous & MCHC & continuous\\
\addlinespace[0.05cm]
BUN & continuous & MCV & continuous\\
\addlinespace[0.05cm]
O2 Sat ($\%$) & continuous & MPV & continuous\\
\addlinespace[0.05cm]
PT-INR & continuous & PT & continuous\\
\addlinespace[0.05cm]
RBC & continuous & PTT & continuous\\
\addlinespace[0.05cm]
RDW & continuous & WBC & continuous\\
\addlinespace[0.05cm]
Alkaline ph. & continuous & Albumin & continuous\\
\addlinespace[0.05cm]
Bedside Glucose & continuous & Anion Gap & continuous\\
\addlinespace[0.05cm]
Calcium & continuous & Bicarbonate & continuous\\
\addlinespace[0.05cm]
Creatinine & continuous & Glucose & continuous\\
\addlinespace[0.05cm]
Lactate & continuous & Magnesium & continuous\\
\addlinespace[0.05cm]
pH & continuous & paCO2 & continuous\\
\addlinespace[0.05cm]
paO2& continuous & Phosphate & continuous\\
\addlinespace[0.05cm]
Platelets & continuous & Potassium & continuous\\
\addlinespace[0.05cm]
Sodium & continuous & Bilirubin & continuous\\
\addlinespace[0.05cm]
Protein & continuous & Troponin - I & continuous\\
\addlinespace[0.05cm]
Urinary s. Gravity & continuous & mean BP & continuous\\
\addlinespace[0.05cm]
SBP & continuous & DBP & continuous\\
  \bottomrule                                 
\end{tabular}
\label{tab:eICU_features}
\end{table*}

\subsection{Code, Benchmark Models, and Training Details}
Sample data and code implementations can be found here: \url{https://anonymous.4open.science/r/DynaGraph-6344}. The requirements.txt file contains the relevant packages and versions.

We coded the implementations of LSTMs, GRUs, Transformer, and TCNs using PyTorch and these can be found in the GitHub repository. We used two layers for all of the models, a dropout of 0.5, 4 heads for the Transformer, and 44 channels with a maximum sequence length of 24 for the TCNs. We used one attention head to match the representative power of the graph models. \\

eICU and MIMIC-III were divided into a training, validation, and test set at an 8:1:1 ratio and using a stratified split per patient based on outcomes. Randomness was controlled for by the setting of seeds throughout the data pre-processing and training stages with the following seeds used: 42, 1992, 1709, 250, 213. \\

All models were trained using the Adam optimizer, and the validation set performance was used to select the final model before the test set results. Batch sizes for benchmark models were 32 for eICU, and 64 for MIMIC-III. DynaGraph was tuned for batch sizes [32, 64, 128] and 128 was found to be optimal. Convergence was sufficient after 100 epochs for eICU, and 150 epochs for MIMIC-III. All models were implemented in PyTorch with an NVIDIA V100 and 50GB of RAM. About 50GB of storage is required to store the data and pre-trained models to run all of the provided notebooks. A full run takes almost 20 minutes on the larger MIMIC III dataset. \\

All models were trained for 100 epochs with early stopping and a learning rate scheduler with patience of 5 epochs. We used an Adam optimiser with a learning rate of 0.001. \\

We implemented grid search hyperparameter tuning on all models with the MIMIC-III dataset. All benchmark models had two layers. Hyperparameters for DynaGraph (Grid search on the validation set) are included with the optimal parameters in bold:
\begin{itemize}
    \item batch size: 32, 64, \textbf{128}
    \item learning rate: 0.01, 0.001, \textbf{0.0001}
    \item groups of graphs or time-slices: \textbf{6}, 8, 12
    \item dropout: \textbf{0.5}, 0.7, 0.9
    \item loss scaling parameters: 0.001 (structural), 0.01 (contrastive), 0.1, 0.5 (regularisation), 1 (the rest)
    \item number of layers in GIN: \textbf{2}, 3, 4
    \item number of layers in MLP: 2, \textbf{4}, 6
\end{itemize}

Hyperparameters for LSTM tuning include batch size (32, 64, \textbf{128}), learning rate (\textbf{0.01}, 0.001, 0.0001), dropout (0.5, \textbf{0.7}, 0.9). For GRU, batch size (32, 64, \textbf{128}), learning rate (\textbf{0.01}, 0.001, 0.0001), dropout (0.5, \textbf{0.7}, 0.9). For Transformer, batch size (32, 64, \textbf{128}), learning rate (0.01, \textbf{0.001}, 0.0001), dropout (0.5, \textbf{0.7}, 0.9). For TCN, batch size (32, 64, \textbf{128}), learning rate (0.01, 0.001, \textbf{0.0001}), dropout (0.5, \textbf{0.7}, 0.9). For GAT, batch size (32, 64, \textbf{128}), learning rate (0.01, 0.001, \textbf{0.0001}), dropout (\textbf{0.5}, 0.7, 0.9). For GCN, batch size (32, 64, \textbf{128}), learning rate (0.01, 0.001, \textbf{0.0001}), dropout (\textbf{0.5}, 0.7, 0.9). For GIN, batch size (32, 64, \textbf{128}), learning rate (0.01, 0.001, \textbf{0.0001}), dropout (\textbf{0.5}, 0.7, 0.9).

\section{Graph Augmentations}

\begin{figure}[t]
    \centering
    \includegraphics[width=0.8\linewidth]{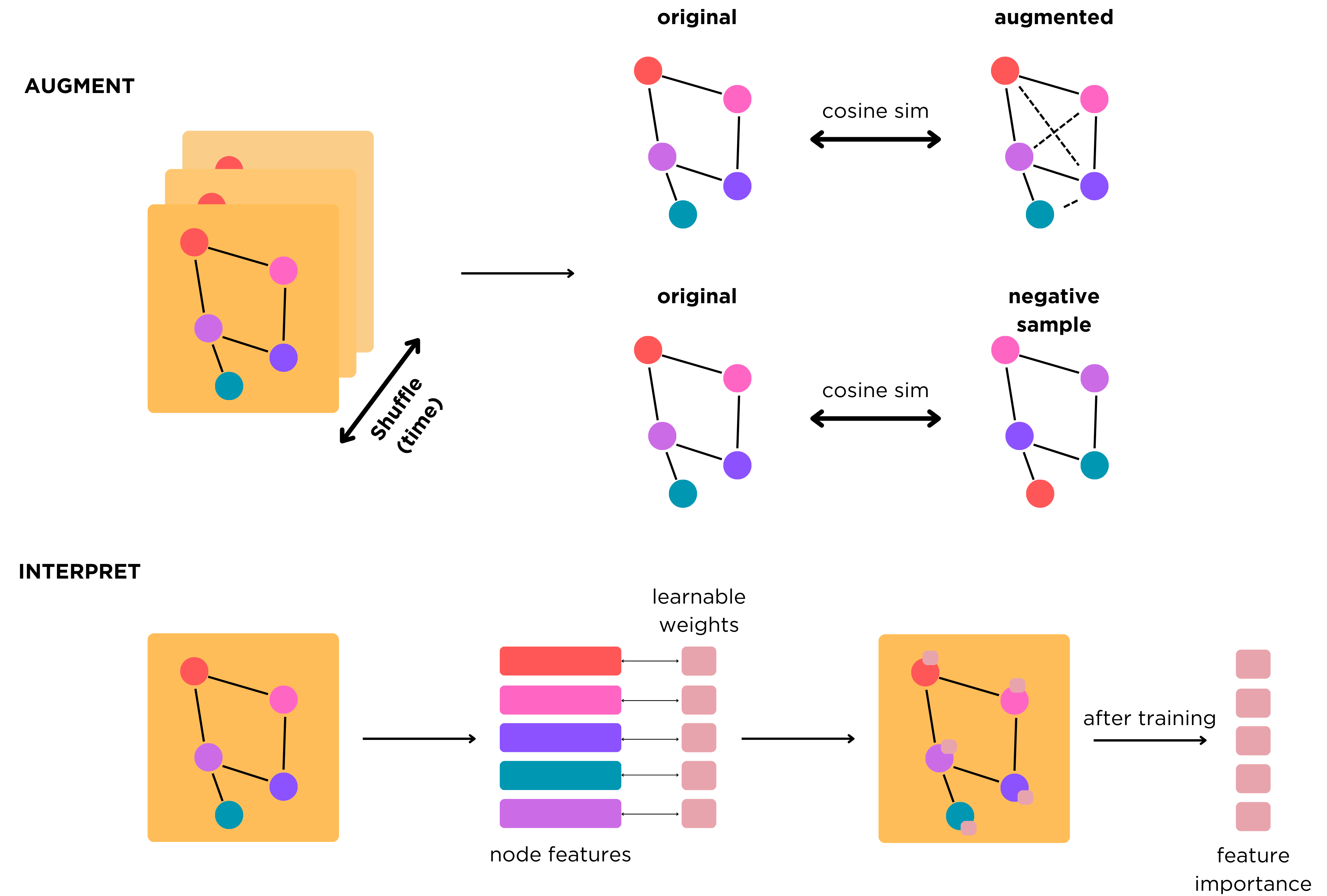}
    \caption{The augmentation and interpretation steps in DynaGraph learning. The dynamic graph constructs represented as adjacency matrices are shuffled along the time axis for added dynamism and noise. This is done per sample. Each graph or adjacency matrix is used to generate an augmented and negative sample pair and measured against either pair using cosine similarity. The similarity score is used in calculating the contrastive loss term. Each original graph will also be paired with an initially uniform interpretability weight matrix where the weights act as indicators for which part of the graph or adjacency matrix contributed the most to the downstream prediction. The weights are a function of the magnitude of the gradient update for that node or edge in the adjacency matrix.}
    \label{fig:Augment}
\end{figure}

A visualisations of the graph augmentations can be seen in Figure \ref{fig:Augment}.

\section{Graph Isomorphism Network (GIN) and Interpretability} \label{GIN}
Graph Isomorphism Networks (GINs) \citep{xu2018powerful} represent a significant advancement in graph neural networks by approximating the Weisfeiler-Lehman (WL) test through its learnable framework. GINs effectively learn powerful node embeddings that capture the topology of the graph, making it possible to distinguish between different graph structures. \\

The update rule for GIN is given by:
\begin{equation}
    h_v^{(k)} = \text{MLP}^{(k)} \left( (1 + \epsilon^{(k)}) h_v^{(k-1)} + \sum_{u \in \mathcal{N}(v)} h_u^{(k-1)} \right)
\end{equation}
where \( h_v^{(k)} \) is the node representation of \( v \) at layer \( k \), \( \mathcal{N}(v) \) denotes the neighbours of \( v \), \( \epsilon^{(k)} \) is a learnable parameter that adjusts the weighting of the node’s own features relative to its neighbours, and \( \text{MLP}^{(k)} \) is a multi-layer perceptron. In DynaGraph, the GIN is used to obtain graph representations at the latent level for mean and variance computation in variational inference for the VGAE as well as a decoder for the final graph output before temporal pooling for downstream tasks. \\

In the context of DynaGraph, we propose a novel approach to enhance the interpretability of predictions in EHR multivariate time-series data. Our model employs a pseudo-attention mechanism that operates on the graph construction phase. This mechanism allows the model to dynamically adjust the graph's structure, thus enhancing the interpretability of the learned representations. The interpretability function for a node \( v \), for example, can be expressed mathematically as:
\begin{equation}
    \text{Score}_v = \sum_{k=1}^{K} \| \text{grad} (y, h_v^{(k)}) \|_2
\end{equation}
where \( y \) is the prediction output, and \( \text{grad} (y, h_v^{(k)}) \) represents the gradient of the loss function at \( y \) with respect to the node embedding \( h_v^{(k)} \). This gradient highlights the influence of each node on the prediction, allowing clinicians and researchers to identify critical features in the graph that significantly impact patient outcomes. \\

Moreover, by integrating this interpretability function directly into the dynamic graph learning process, DynaGraph not only adapts to changes in data over time but also provides insights into how these changes influence predictive outcomes. This method bridges the gap between complex model predictions and clinical decision-making by providing a transparent view of feature importance over time. \\

Through this approach, DynaGraph leverages the strengths of GIN in a dynamic setting while also addressing the crucial need for interpretability in models used in healthcare. This dual focus on performance and transparency is especially vital in clinical applications where understanding the reasoning behind model predictions can significantly impact patient care and treatment strategies.

\section{Convergence Proof for the Interpretability Matrix in DynaGraph}
\label{Interpretability}
\subsection*{Definitions and Assumptions}
Let \( G_t = (V, E_t, W_t) \) denote the graph at epoch \( t \), where:
\begin{itemize}
    \item \( V \) is the set of nodes.
    \item \( E_t \) is the set of edges.
    \item \( W_t \) are the node and edge interpretability weights from $I_s$ for a time slice $s$ at epoch \( t \).
\end{itemize}

\textbf{Convergence Definition:} We define convergence of the interpretability matrix in terms of the weights:
\[ \lim_{t \to \infty} \|W_t - W_{t-1}\|_F = 0, \]
where \( \| \cdot \|_F \) denotes the Frobenius norm. \\

\textbf{Assumptions:}
\begin{enumerate}
    \item The weights \( W_t \) are updated by an optimization algorithm aimed at minimizing a loss function \( \mathcal{L} \), which is differentiable and has Lipschitz continuous gradients.
    \item The learning rate \( \eta \) used in the updates satisfies \( \eta_t \to 0 \) as \( t \to \infty \), and \( \sum_{t=1}^\infty \eta_t = \infty \), \( \sum_{t=1}^\infty \eta_t^2 < \infty \).
\end{enumerate}

\subsection*{Proof of Convergence}
Consider the gradient update rule for the weights:
\[ W_t = W_{t-1} - \eta_t \nabla_{W} \mathcal{L}(W_{t-1}). \]

Using the Taylor expansion of \( \mathcal{L} \) around \( W_{t-1} \), we have:

\[ \mathcal{L}(W_t) \approx \mathcal{L}(W_{t-1}) + \nabla_{W} \mathcal{L}(W_{t-1})^\top (W_t - W_{t-1}) + \] 
\[\frac{1}{2} (W_t - W_{t-1})^\top H (W_t - W_{t-1}), \]
where \( H \) is the Hessian matrix of \( \mathcal{L} \).

Substituting the update rule, we get:
\[ \mathcal{L}(W_t) \approx \mathcal{L}(W_{t-1}) - \eta_t \|\nabla_{W} \mathcal{L}(W_{t-1})\|_F^2 + \]
\[\frac{1}{2} \eta_t^2 \|\nabla_{W} \mathcal{L}(W_{t-1})\|_F^2 \|H\|_F. \]

Since \( \eta_t \to 0 \) and assuming \( \|H\|_F \) is bounded, the second-order term vanishes faster than the first-order term. Therefore, the change in the loss function due to weight updates diminishes over time, implying:
\[ \lim_{t \to \infty} \|\nabla_{W} \mathcal{L}(W_{t-1})\|_F = 0, \]
and thus:
\[ \lim_{t \to \infty} \|W_t - W_{t-1}\|_F = 0. \]

\textbf{Conclusion:} Under the assumptions of Lipschitz continuity of gradients and appropriate conditions on the learning rate, the explainable graph's weights converge, indicating stability and reliability in the graph-based explanations generated by the model.

\section{VGAE} \label{VGAE}
Continuing from the notation in the Methods section, the aggregation of the different components from the adjacency, interpretability, and embeddings matrices is performed as:
\begin{equation}
    G^{(i)} = \left( A^{(i)} \parallel I^{(i)} \parallel E^{(i)} \right)
\end{equation}
where $\parallel$ denotes the concatenation operation along the feature dimension for a sample $i$. 

GIN operates by transforming the dynamic graph data through a series of graph isomorphism layers that adapt to temporal shifts in the graph structure. The encoder within the VGAE is formulated to compute the parameters of the latent variables as:
\begin{equation}
    \mu^{(i)}, \log \sigma^{2(i)} = \text{GIN}_{\text{encoder}}(G^{(i)})
\end{equation}
These parameters facilitate sampling of the latent representation $z^{(i)}$ using the reparameterization trick:
\begin{equation}
    z^{(i)} = \mu^{(i)} + \sigma^{(i)} \odot \epsilon, \quad \epsilon \sim \mathcal{N}(0, I)
\end{equation}
where $\odot$ denotes element-wise multiplication. The decoder part of the VGAE attempts to reconstruct the adjacency matrix from the latent representations, thereby capturing the evolving graph structure:
\begin{equation}
    \hat{A}^{(i)} = \sigma(\text{GIN}_{\text{decoder}}(z^{(i)}))
\end{equation}
where $\sigma$ is the sigmoid activation function, converting output values to probabilities. The loss function for the VGAE incorporates the reconstruction error of the adjacency matrix and the Kullback-Leibler divergence, promoting efficient graph structure learning:
\begin{equation}
    \mathcal{L}_{\text{VGAE}} = \text{BCE}(A^{(i)}, \hat{A}^{(i)}) + \text{KL}(\mathcal{N}(\mu^{(i)}, \sigma^{2(i)}) \| \mathcal{N}(0, I))
\end{equation}
where BCE represents the binary cross-entropy loss, and KL is the Kullback-Leibler divergence. Details on temporal pooling which follows after the VGAE reconstructs the graph can be found in Appendix section \ref{tgp}. The pooled graph is then flattened as input into a multi-layer perceptron for downstream task classification, ie. multi-label classification.

\section{Temporal Pooling}
\label{tgp}
The output of the VGAE module is a complex graph representation which to be used in the downstream prediction task can be pooled. Usual approaches would flatten this representation and thus lose important temporal patterns that could be useful for the prediction task. Taking inspiration from \citep{liu2023todynet}, we use a 2-dimensional convolutional neural network to cluster the nodes in these graph representations thereby reducing the dimensionality of the problem while preserving information given by VGAE. The convolutional weights are then used to reconstruct the adjacency matrices for the lower-level representations. A 2D CNN layer is designed given the pooled ratio and nodes are considered as the channels to extract features. \\

Given an input node embedding $X_l$ at $l$-th layer (note this $l$ is different from the previous one for time-steps), the $\mathrm{CNN}$ gives an output embedding $X_{l+1}$ as:
$
X^{l+1}=\sum_{j=0}^{N^l-1} \operatorname{W}\left(N^{l+1}, j\right) \star X^l+\operatorname{b}\left(N^{l+1}\right)
$
where $W$ is the convolutional weight, $b$ is the bias term, $\star$ is the cross-correlation operator, $N^l$ denotes the input nodes and $N^{l+1}$ the output nodes for $l$-th layer. \\

We use the output tensor $X^{l+1}$ to then compute the corresponding adjacency matrix of a lower-dimensional graph. Assuming the shape of the learnable weights $W^l$ is $\left[N^{l+1}, N^l, 1\right.$, $kernel size$], then the vector of the learnable parameters in layer $l$ is $V^l \in \mathbb{R}^{1 \times k}$. Define a learnable matrix $M^l=W^l \cdot V^l \in \mathbb{R}^{N^{l+1}} \times N^l$ with rows corresponding to $N^{l+1}$ nodes or clusters and columns corresponding to $N^l$ clusters. Given matrix $M$ and adjacency matrix $A^{l}$ for input data in layer $l$, the output adjacency matrix $A^{l+1}$ is:

$
A^{l+1}=M^{l} A^{l} M^{l^T} \in \mathbb{R}^{N_{l+1} \times N_{l+1}}
$

In summary, $X^{l+1}$ represents the output cluster embeddings after aggregating input embeddings and $A^{l+1}$ denotes the connections and weights of the new clusters. Each element $A_{i j}^{l+1}$ represents the edge weight between $i$ and $j$, thus ensuring a graph representation. This pooling approach is hierarchical and differentiable while preserving temporal information and optimising the clusters during training. \\

\bibliography{main}